\documentclass[preprint,3p,onecolumn,authoryear]{elsarticle}
\usepackage{amssymb}
\usepackage{amsmath}
\journal{Neural Networks}

\usepackage{ragged2e}
\usepackage{float}
\usepackage{xcolor}
\usepackage{xspace}
\usepackage{booktabs}
\usepackage{makecell}
\usepackage{multirow}
\usepackage{verbatim}
\usepackage{subcaption}
\usepackage[colorlinks=true,linkcolor=blue]{hyperref}
\usepackage[linesnumbered,ruled,vlined]{algorithm2e}
\usepackage{enumerate}
\usepackage[shortlabels]{enumitem}

\def\proposal{EvoPruneDeepTL\xspace}
\def\Proposal{Evolutionary Pruning for Deep Transfer Learning\xspace}

\begin{document}
\let\WriteBookmarks\relax
\def\floatpagepagefraction{1}
\def\textpagefraction{.001}

 \newenvironment{idea}{}{}
 %{\color{blue}}{}
 
 \newenvironment{idea2}{}{}
\begin{frontmatter}
\title{EvoPruneDeepTL: An Evolutionary Pruning Model for Transfer Learning based Deep Neural Networks}


%\tnotetext[2]{The second title footnote which is a longer text matter
%   to fill through the whole text width and overflow into
%   another line in the footnotes area of the first page.}


\author[1]{Javier Poyatos}
\ead{jpoyatosamador@ugr.es}

\author[1]{Daniel Molina\corref{mycorrespondingauthor}}
\cortext[mycorrespondingauthor]{Corresponding author}
% \cormark[1]
\ead{dmolina@decsai.ugr.es}

\author[2]{Aritz D. Martinez}
\ead{aritz.martinez@tecnalia.com}

%\author[1]{Siham Tabik}
%\ead{siham@ugr.es}

\author[2,3]{Javier Del Ser}
% \cormark[2]
% \fnmark[1,3]
\ead{javier.delser@tecnalia.com}

\author[1,4]{Francisco Herrera}
% \cormark[2]
% \fnmark[1,3]
\ead{herrera@decsai.ugr.es}

\affiliation[1]{organization={Department   of   Computer   Science   and   Artificial   Intelligence,   Andalusian   Research   Institute   in   Data   Science   and   Computational  Intelligence  (DaSCI),  University  of  Granada},
    %addressline={Radarweg 29}, 
    city={Granada},
    % citysep={}, % Uncomment if no comma needed between city and postcode
    postcode={18071}, 
    country={Spain}}

\affiliation[2]{organization={TECNALIA, Basque Research \& Technology Alliance (BRTA)},
    city={Derio},
    % citysep={}, % Uncomment if no comma needed between city and postcode
    postcode={48160}, 
    %state={Trivandrum},
    country={Spain}}

\affiliation[3]{organization={University of the Basque Country (UPV/EHU)},
    %addressline={Mepukada}, 
    city={Bilbao},
    % citysep={}, % Uncomment if no comma needed between city and postcode
    postcode={48013}, 
    %state={Trivandrum},
    country={Spain}}

\affiliation[4]{organization={Faculty of Computing and Information Technology, King Abdulaziz University},
    %addressline={Mepukada}, 
    city={Jeddah},
    % citysep={}, % Uncomment if no comma needed between city and postcode
    postcode={21589}, 
    %state={Trivandrum},
    country={Saudi Arabia}}

% Footnote of the first authorIn
%\credit{Conceptualization of this study, Methodology, Software}
%\credit{Data curation, Writing - Original draft preparation}

%\cortext[cor2]{Principal corresponding author}

\begin{abstract}
    \justifying \indent
  In recent years, Deep Learning models have shown a great performance in complex optimization problems. They generally require large training datasets, which is a limitation in most practical cases. Transfer learning allows importing the first layers of a pre-trained architecture and connecting them to fully-connected layers to adapt them to a new problem. Consequently, the configuration of the these layers becomes crucial for the performance of the model. Unfortunately, the optimization of these models is usually a computationally demanding task. One strategy to optimize Deep Learning models is the pruning scheme. Pruning methods are focused on reducing the complexity of the network, assuming an expected performance penalty of the model once pruned. However, the pruning could potentially be used to improve the performance, using an optimization algorithm to identify and eventually remove unnecessary connections among neurons. This work proposes \proposal, an evolutionary pruning model for Transfer Learning based Deep Neural Networks which replaces the last fully-connected layers with sparse layers optimized by a genetic algorithm. Depending on its solution encoding strategy, our proposed model can either perform optimized pruning or feature selection over the densely connected part of the neural network. We carry out different experiments with several datasets to assess the benefits of our proposal. Results show the contribution of \proposal and feature selection to the overall computational efficiency of the network as a result of the optimization process. In particular, the accuracy is improved, reducing at the same time the number of active neurons in the final layers.
\end{abstract}
% \includegraphics{grabs.pdf}

%\item Research highlights item 1
%\item Research highlights item 2
%\item Research highlights item 3

% Keywords

\end{frontmatter}

\newpage

\section{Introduction} \label{sec:introduction} 

Deep Learning (DL) \citep{Goodfellow-et-al-2016} is one of the most attractive research areas in machine learning in recent times, due to the great results offered by such models in a plethora of applications. DL architectures are successfully used in many problems, like audio classification \citep{NIPS2009_3674}, audio recognition \citep{noda2015audio}, object detection \citep{zhou2017application}, image classification for medical analysis \citep{DLmedical} or vehicular perception \citep{DLcar}.

Convolutional Neural Networks (CNNs) \citep{lecuncnn} constitute the state-of-the art in image classification \citep{cnnim}. CNNs include two parts, the first part is actually a feature extractor based on convolution and pooling operations. The second part usually contains one or more fully connected layers. In these fully-connected layers, the neuron of each layer is connected to all the neurons of the previous layer, which generates a large number of weights to be trained. The design of an appropriate network for each problem is a requirement in order to obtain a good performance. The training process of a DL architecture is frequently time-consuming. Complexity reduction maintaining the performance is an important challenge in DL, currently attracting significant attention in the community. Transfer Learning (TL) \citep{Weiss2016} is usually considered the alternative. It is very common to use a DL model with fixed and pre-trained weights in the convolutional layers  with a dataset (like ImageNet \citep{cnnimagenet}) and then add and train several layers, named fully-connected layers, to adapt the network to a different classification problem \citep{shinDeepConvolutionalNeural2016,khan2019novel,SRSMAS}.

% parrafo situando el debate del problema, la optimizacion de las capas, y mencionando el pruning como potencial solucion y la reduccion de la complejidad
The architecture of fully-connected layers used for the problem is a critical decision, and its design is still an open issue in terms of the number of layers and neurons per layer \citep{liuSurveyDeepNeural2017}. There are general guidelines based on the experience working with these layers, rather than rules to follow for the configuration of them. Therefore, any kind of optimization in them could provide a benefit in terms of model complexity and performance. The pruning approaches follow the key idea of reducing the complexity of the model, which creates new networks with less computational cost for training. This idea is followed in \citep{frankle2019lottery}, which also shows that, in the end, the accuracy can also improve as a result of pruning.

Pruning is interpreted as removing unnecessary connections from the model, but learning which connections are the fittest to improve the performance of the model is the key point. In fact, the selection of the best \textit{features} for the problem is known as Feature Selection (FS) \citep{Iguyon20031157}. In our case, TL allows the extraction of the features of the input data of the DL model. These features are the input of the fully-connected layers that will be trained and, as a result of that, will largely affect the performance of the network. Nonetheless, in many cases, the problem that is formulated to learn these features is usually different, sometimes more complex, than the one at hand and, therefore, not all the learned patterns would be required. For that reason, FS gives rise to an interesting option to select and retain the subset of all features that lead to an improved performance of the model \citep{yildirimFS}.

%Evolutionary Algorithms \citep{EABook} (EAs) are optimization algorithms. They have been successfully applied to many complex optimization problems. Even though they cannot guarantee the achievement of the optimum for the problem at hand, they obtain good results with limited resources and reasonable processing time. Another advantage is their versatility: several of them, like genetic algorithms \citep{GA} (GAs) allow optimizing solutions with different representations \citep{GABook}. The spectrum of problems in which EAs can be used is very wide. 

% challenge 

In pruning scenarios, the main aim of most of the traditional pruning techniques mainly aim at reducing the number of trainable parameters of the network, at the cost of a lower performance. They seek to control the performance degradation resulting from the process, but it is not their priority. Furthermore, they locally optimize parts of the network rather than searching for globally optimal pruning policies, yielding usually suboptimal pruned subnetworks with a lower performance. Another disadvantage of these pruning proposals is the fact that, as the pruning affect all layers, the complete network must be trained again, hence obtaining no advantages from the TL process. It could be useful to have a pruning technique that prioritizes results over complexity reduction, targeting a global performance improvement of the network while reducing its complexity.

% hypothesis -> los algoritmos evolutivos nos permiten mejorar a las técnicas de pruning debido a que somos capaces de mejorar la precisión de los modelos mientras que reducimos la complejidad de la red.

Transforming the fully-connected layers into a sparse representation, in which each connection could be active or inactive, could be used to prune neural networks. Following this approach, both pruning and FS can be seen as optimization problems, in which the target is to obtain the active set of connections that produce the best performance. This optimization problem can be globally tackled by optimization algorithms like Evolutionary Algorithms \citep{EABook} (EAs). They have been successfully applied to many complex optimization problems. Even though they cannot guarantee the achievement of the optimum for the problem at hand, they obtain good results with limited resources and reasonable processing time. Another advantage is their versatility: several of them, like genetic algorithms \citep{GA} (GAs) allow optimizing solutions with different representations \citep{GABook}. The spectrum of problems in which EAs can be used is very wide. EAs have been traditionally applied to optimize neural networks \citep{ibaEvolutionaryApproachMachine2018a}, but their usage in DL networks to improve DL networks \citep{MARTINEZ2021161}, to train them \citep{adaswarmcoello}, and to create new DL networks from scratch \citep{elsken2019neural} is more recent. The use of EA's is mainly oriented towards optimizing a complete network. However, in this paper, our aim is to adapt the fully-connected layers (the only trained for the problem to solve using TL) to improve the accuracy in the predictions, together with the complexity reduction. Our main hypothesis is the convenience of use of EAs to prune the fully-connected layers via a sparse representation. 

% proposal

We propose an evolutionary pruning model based on TL for deep neural networks, \Proposal (\proposal). \proposal can be applied to a DL model that resorts to TL to tackle a new task. \proposal combines sparse layers and EA, consequently, neurons in such layers are pruned to adapt their sparsity pattern to the addressed problem. \proposal is able to efficiently explore the neuron search space (to discover coarsely grained solutions) or, alternatively, in the connection search domain (fine-grained solutions). 

An important aspect to analyze in \proposal is that one of its solution encoding schemes effectively leads to a feature selection mechanism, in which we deactivate the extracted features and the EA evolves these features to learn which ones fit best as predictors for the given problem. 

\begin{idea}\proposal' goals include flexibility and adaptability. \proposal has been designed to be flexible, and the automatic configuration of the network can be applied to different pre-trained networks, used as feature extractors, and different fully-connected layers. This make our proposal capable of tackling different problems. The \begin{idea2}removal of connections\end{idea2} that \proposal \begin{idea2}performs\end{idea2} allows the model to be adaptable to the specific dataset to be modeled. Thus, when the dataset suffers a change, the resultant configuration will also be adapted to the new circumstances.
\end{idea}

% experimental study

To assess the performance of \proposal, we have conducted an extensive experimentation that leads to several valuable insights. To begin with, experimental results showcase the behavior and effectiveness of \proposal in terms of precision and in terms of reduction of the complexity of the network. Thanks to the flexibility of \proposal, it is applied to perform either pruning or FS. Both cases improve the accuracy of the network when the comparison is made against reference models and CNN pruning methods from the literature. Moreover, in most cases, the FS scheme achieves a better performance than the pruning scheme in terms of the accuracy of the network. Furthermore, the network pruned by the FS scheme also achieves a significantly reduced number of connections in its fully connected part, contributing to the computational efficiency of the network. \begin{idea}We have also included several experiments showing the flexibility of the model, both changing the feature extractor and showing how changes in the dataset implies a modification in the final configuration obtained by \proposal.\end{idea} In short, this extensive experimentation is used to provide answer to the following \begin{idea}six\end{idea} questions as the thread running through this experimental study:

\begin{enumerate}[start=1,label={(\bfseries RQ\arabic*)}]
    \item Which is the performance of \proposal against fully-connected models? 
    \item Which would be better, to remove neurons or connections?
    \item Which is the performance of \proposal when compared to other efficient pruning methods?
    \item Which would be better, the use of pruning of fully-connected layers or Feature Selection?
    \begin{idea}
    \item How does \proposal perform when applied to different pre-trained networks?
    \item Can \proposal adapt efficiently their pruned knowledge to changes in the modeling task, showing robustness?
    \end{idea}
\end{enumerate}

% paper description

The rest of the article is structured as follows: Section \ref{sec:relatedwork} exposes related work to our proposal present in the literature. Section \ref{sec:proposal} shows the details of the proposed \proposal model. Section \ref{sec:framework} presents our experimental framework. In Section \ref{sec:results}, we show and discuss the \proposal's results of the experiments of pruning, feature selection and against efficient CNN pruning methods of the literature. \begin{idea}Moreover, \proposal is tested with different extractor features and with different variations of datasets in this section. Section \ref{sec:lessons} follows by summarizing the advantages and drawbacks of our proposal when compared to other pruning approaches. \end{idea} Finally, Section \ref{sec:conclusions} draws the main conclusions stemming from our work, and outlines future work departing from our findings.

\section{Related work}\label{sec:relatedwork}

The purpose of this section is to make a brief review of contributions to the literature that link to the key elements of our study: Transfer Learning (Subsection \ref{sec:tf}), Neural Architecture Search (Subsection \ref{sec:nas}), CNN pruning (Subsection \ref{sec:relatedwork-pruning}), Evolutionary Algorithms (Subsection \ref{sec:eacnn})  and Feature Selection with Deep Learning (Subsection \ref{sec:fsdl}).
\subsection{Transfer Learning} \label{sec:tf}

TL \citep{panSurveyTransferLearning2010} is a DL mechanism encompassing a broad family of techniques \citep{tan2018survey}. Arguably, the most straightforward method when dealing with neural networks is \textit{Network-based deep transfer learning}, in which a previous network structure with pre-trained parameters in a similar problem is used. It offers good results by the behavior of DL models, in which first layers detect useful features on the images, and later layers strongly depend on the chosen dataset and task. As finding these standard features on the first layers seems very common regardless of the natural image datasets, its trained values can be used for different problems \citep{yosinskiHowTransferableAre2014a}. Training DL models from scratch is usually time-consuming due to the great amount of data in most cases. TL gives some benefits which make it a good option for DL: reduction of time needed for training \citep{sa2016deepfruits}, better performance of the model and less need of data.

TL has been applied to several real-world applications, such as sound detection \citep{jung2019polyphonic} or coral reef classification \citep{GOMEZRIOS2019315}. Moreover, in \citep{transfer2016} two different approaches for TL are discussed: fine-tuning or full training. They demonstrated that, for medical reasons, a pre-trained CNN with adequate fine-tuning performed better in terms of accuracy than a CNN trained from scratch. Another approach of TL is presented in \citep{MEHDIPOURGHAZI2017228}, in which an optimization of TL parameters for plant identification is proposed.

% en este parrafo incluimos propuestas que tengan resnet y TL a ser posible o bien el uso de resnet para problemas.
There are different deep neural networks proposed in the literature. One of the most popular is ResNet, which uses residual learning to improve the training process, obtaining better performance than other models \citep{He2016770}. ResNet models are characterized by the use of deeper neural networks without loss of information due to their architecture. Different ResNet models with TL have been used in several applications \citep{resnettransfer}, such as medical classification like pulmonary nodule \citep{nibali2017pulmonary} and diabetic retinopathy classification\citep{Wan2018274}. \begin{idea}
Moreover, other networks have shown great performance when used with TL, such as DenseNet \citep{densenet} and VGG \citep{vgg}. An example of DenseNet with TL is presented in \citep{densenettl}, which shows that this network architecture is able to achieve a great result for the task at hand when combined with TL. Lastly, VGG has also shown an outstanding performance when it is used in combination with TL. An example is presented in \citep{vggtl} in which a pre-trained VGG-19 network is used to solve a fault diagnosis problem. \end{idea}

\subsection{Neural Architecture Search} \label{sec:nas}

The appropriate design of a neural network is a key point to solve DL problems. Nevertheless, finding the best architecture that optimally fits the data and, as a result of that, gives the best outcome for the problem is extremely difficult. Recently, the term Neural Architecture Search, NAS, has obtained a great importance in this field. The objective of NAS is the automatic search for the best design of a NN to solve the problem at hand.

The first work in this field is presented in \citep{Stanley200299} in the beginning of this century, in which they demonstrate the effectiveness of a GA to evolve topologies of NN. 

In \citep{Zoph20188697}, the authors design the NASNet architecture, a new search space to look for the best topology for the tackled problem. Moreover, in \citep{Liu2018ProgressiveNA} the authors propose to search for structures in increasing order of their level of complexity, while learning a surrogate model to guide the search through structure space. 

NAS methods usually rely on Reinforcement Learning, RL, and EA, like \citep{Zoph20188697} or \citep{Liu2018}, in which the authors explore the search space using a hierarchical genetic representation. Another example of RL for NAS is shown in \citep{Kokiopoulou2019FastTA}. The authors propose a novel method that, by sharing information on multiple tasks, is able to efficiently search for architectures.

NAS can also be viewed as a multi-objective problem. Among these methods, one of them is presented in \citep{Elsken2019EfficientMN}, in which the authors propose a multi-objective for NAS that allows approximating the entire Pareto-front of architectures. Another example is Neural Architecture Transfer \citep{NAT}, that allows to overcome a common limitation of NAS, that is requiring one complete search for each deployment specification of hardware or objective. They use an integrated online transfer learning and a many-objective evolutionary search procedure.

Recently, one of the most well-known multi-objective EA, NSGA-II, has been used for NAS \citep{Lu2019419}, called NSGA-Net. This novel proposal looks for the best architecture through a three-step search: an initialization step from hand-crafted architectures, an exploration step that performs the EA operators to create new architectures, and finally an exploitation step that utilizes the knowledge stored in the history of all the evaluated architectures in the form of a Bayesian Network.

Lastly, there are more advanced techniques of NAS and EA given by \citep{real2019regularized}, in which a new model for evolving a classifier is presented, and by \citep{real2020automl}, in which the authors propose AutoML-Zero, an evolutionary search to build a model from scratch (with low-level primitives for feature combination and neuron training) which is able to get a great performance over the addressed problem.

\subsection{CNN Pruning}\label{sec:relatedwork-pruning}

The main reason to optimize the architecture of a deep neural network is to reduce its complexity. That reduction can be done in different ways \citep{Long2019}. One of them is by designing compact models from scratch instead of resorting to architectures comprising multiple layers. Another strategy is via weights-sharing \citep{Ullrich2017}. An alternative method to reduce the complexity of DL models is low-rank factorization \citep{Long201951866}, based on a matrix decomposition to convolutional layers to estimate parameters. However, one of the most popular is Network Pruning. The objective of pruning is to remove unnecessary parameters from a neural network, so that they do not participate during training and/or inference. It can be done in the convolutional phase on the channels, kernels and weights or even in the fully connected phase on the neurons. In \citep{masson2019survey} they show a classification of pruning methods for channels. They categorize the pruning methods for channel reduction, and they also specify the criteria used to select these channels: based on weights or based on feature maps.

We have seen that network pruning has achieved a great importance in the literature as many researchers have applied different techniques to simplify a CNN using a pruning scheme. In \citep{liu2018rethinking} they classify the pruning methods in unstructured and structured pruning, and make a review of all the state-of-art structured pruning methods. Unstructured pruning methods remove weights without following any order. For the structured methods, there are some rules or even constraints which define how the pruning is done \citep{Anwar2017}. Typically, the pruned layers appertain to the convolutional phase \citep{Luo20175068}. In our proposal, we instead apply a structured pruning scheme to the fully-connected layers.

Among pruning methods, the value-based \textit{weight pruning} \citep{han2015learning} and \textit{neuron pruning} \citep{srinivas2015data} have arisen as the most used, particularly due to their simplicity. The logic behind this pruning methods is straightforward: a certain amount (\%) of the weights or neurons that contribute less to the final trained model are removed from the architecture. This makes the network quicker to perform inference and endows it with better generalization capabilities. However, multiple pruning and retraining steps demonstrated that it is possible to recover fully or partially the knowledge lost in the pruning phase. Further along the series of pruning approaches published to date, \textit{Polynomial Decay} \citep{zhu2017prune} is a scheduled pruning method that considers that a higher amount of weights can be pruned in early stages of pruning, while systematically less amount of weights should be pruned in late stages. Between pruning steps, the network is retrained for some epochs. An implementation of the discussed methods can be found for Tensorflow. \footnote{https://www.tensorflow.org/model\_optimization/guide/pruning, Tensorflow Pruning. Last access: $28/01/2022$}

Pruning a CNN model reduces its complexity, but sometimes leads to a decrease of the performance of the model, although there are some proposals that reduce the complexity of the model with no loss of accuracy \citep{Han2016}. 

% algoritmos geneticos y DL: propuestas de geneticos + DL % revisar
Pruning a neural network can be conceived as an optimization method in which we start from the original vector, and connections/neurons are decision variables whose value is evolved towards optimizing a given objective. In this context of evolution of neural networks, evolutionary algorithms for evolving DL architectures have been applied \citep{ibaEvolutionaryApproachMachine2018a}. While this combination of EA and DL models seemed to be a great scheme for the optimization of DL models, especially for CNN network, the optimization of DL models is still an open problem \citep{liuSurveyDeepNeural2017}. Many proposals have been published about this problem like in \citep{MARTINEZ2021161}, where they make a review of proposals using EAs for optimizing DL models, prescribing challenges and future trends to effectively leverage the synergy between these two areas.

Researchers have presented a great variety of proposals about the optimization of DL models using EA, most commonly for CNN. In \citep{Comput2018}, the authors developed EvoDeep, an EA with specific mutation and crossover operators to automatically create DL models from scratch. Moreover, in \citep{Real2017a} a novel evolution approach to evolve CNN models using a GA was proposed. Another example of the optimization of CNN was developed in \citep{assunccao2019denser}, in which a GA was presented for the optimization of the topology and parameters of the CNN. 

In our proposal, we improve the performance of the models using a TL approach to extract the features of the images and apply a reduction of the fully-connected layer using a GA to optimize a sparse layer.

\subsection{Evolutionary Algorithms for CNN Pruning} \label{sec:eacnn}

In the previous section, several works of CNN pruning have been presented, but none of them use an EA to prune. In this section, we mention some studies present in the literature which have used an EA in order to prune a CNN model. To begin with, in \citep{Liu20172245}, the authors propose a sparse approach to reduce CNN complexity. EAs are also a good way to prune CNN. In \citep{MANTZARIS2011831}, a first attempt of pruning and GA is proposed for a medical application. They use a GA to search for redundancy factors in a neural network. Moreover, in \citep{Samala2018} another EA is presented to prune deep CNN for breast cancer diagnosis in digital breast tomosynthesis. A combined approach of EA and sparse is proposed by Wang et al. \citep{WANG2020247}, in which a GA and sparse learning are applied to a scheme of network channel pruning in the convolutional scheme of the CNN. For pruning CNN, not only GAs but also other algorithms are used, like Differential Evolution (DE). In \citep{salehinejad2020edropout} the authors propose to use a Differential Evolution algorithm to prune the convolutional phase and the fully-connected phase of some Deep CNN, obtaining a reduction of the model but a small decrease of its performance. 

However, all previous works are focused on reducing the complexity, using the EA to reduce the accuracy loss of the pruned network. Also, many of them try to reduce the whole model, changing the complete architecture and making the pre-trained values unusable. The re-training of the network may be a time-consuming task, so we assume that TL is useful in this context. We therefore maintain the original architecture with pre-trained values. Our model focuses on improving the performance of the model by pruning connections of the fully-connected layers using a GA to evolve the connections. In this environment, the search space of the GA is narrower and a faster convergence of the algorithm may be reached.

In addition to that, in the field of neural architecture search, more advanced techniques have been developed. Among them, in Section \ref{sec:nas}, either \citep{real2019regularized} and \citep{real2020automl} have been commented. Nonetheless, they also have a great relevance in this section. The first one evolves a classifier, whereas in the second one, the authors develop an evolutionary search to build a model from scratch.

\subsection{Feature Selection and Deep Learning} \label{sec:fsdl}

One of the advantages of using TL is reducing the required time to train a DL architecture. Nonetheless, the result of this process may lead to recognize patterns that are not useful to address the problem at hand. For that reason, once TL is applied, a FS process to obtain the best features might lead to a better performance of the neural network \citep{fsdnn}.

An example of this process is presented in an arrhythmia detection task addressed in \citep{yildirimFS}, in which the authors propose a mechanism based on feature extraction and selection to improve and ultimately obtain one of the best results for this problem. In relation to medical problems, the combination of FS and DL is also used in cancer diagnosis and digital breast tomosynthesis. In \citep{Samala2018} they use a TL approach and then a FS process followed by an evolution through a GA that leads to a reduced network with the same performance. Another example is described for remote sensing scene classification, in which the FS makes an impact to improve the performance of the neural network models \citep{ZouFS}, as the authors formulate the FS problem as a feature reconstruction problem. Their iterative method selects the best features to solve this problem as the discriminative features.

In our proposal, if we assume that TL is applied and we only have one fully-connected layer, then the pruning is made in relation to the extracted features of the network and, therefore, we are making a selection of the features that adjust at best to the tackled problem. 

\section{\Proposal} \label{sec:proposal}

This section describes \proposal, which is a model that replaces fully-connected layers with sparse layers \begin{idea2}which are being evolved\end{idea2} using a genetic algorithm in a TL approach. Subsection \ref{sec:gs} gives a notion of the concept of sparse layer and the description of \proposal. In Subsection \ref{sec:ga-elements}, we define the evolutionary components of \proposal. The description of the process of creating the network and how the pruning is made is shown in Subsection \ref{sec:network}.

\subsection{Global scheme of \Proposal} \label{sec:gs}

In a fully-connected layer, all neurons of each side are connected. Sometimes, all these connections may not be necessary, and the learning process can be reduced. For that reason, the fully-connected layer can be replaced by a sparse layer, in which some connections are eliminated.

In this work, our goal is to improve the performance of the neural network and, at the same time, to decrease the maximum number of connections or neurons. To this end, we use a sparse layer, which is composed of a subset of all connections of a fully-connected layer.

\begin{figure}[H]
    \centering
    \subfloat[\centering Fully-Connected Layer]{{\includegraphics[width=2.8cm]{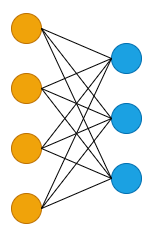}}
    \label{fig:fullybuena}}
    \qquad
    \subfloat[\centering Sparse Layer]{{\includegraphics[width=2.8cm]{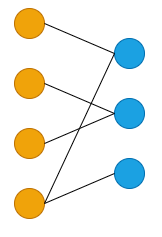} }
    \label{fig:sparsebuena}}
    \caption{Representation of both architectures}
    \label{fig:sparse}
\end{figure}

Fig. \ref{fig:fullybuena} shows the fully-connected network architecture, while Fig. \ref{fig:sparsebuena} represents the sparsely connected architecture with a connection matrix of $4 \times 3$ because we have 3 classes (blue circles) and 4 neurons of the previous layer.

\begin{idea}

In this section, we discuss the basic notions of pruning and sparse layers. Moreover, the encoding strategies of \proposal are described, together with the decoding process of the chromosome encoding the pruning pattern (genotype) that yields the pruned sparse layer(s) (phenotype). Nonetheless, for the sake of a clear vision of \proposal, Fig. \ref{fig:evoprunedeeptl} shows a diagram that exposes its general components. We next complement the detailed description provided in the following subsections with a short, albeit illustrative introduction to the key parts and overall workflow of \proposal:

\begin{figure}[htp]
    \centering
    \includegraphics{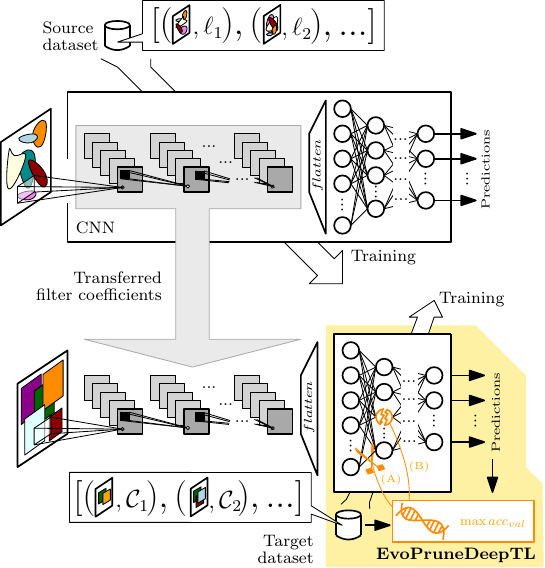}
    \caption{Diagram of \proposal.}
    \label{fig:evoprunedeeptl}
\end{figure}

First, the TL process departs from a source dataset modeled by a CNN model, which adjusts the value of its coefficients by means of its learning algorithm. Then, those parameters are transferred to another network aimed to model a target task. This implicitly assumes that both tasks are correlated with each other, such that the knowledge delivered from the source to the target task via the transferred network weights can positively contribute to the learning process of the target task. These weights are kept fixed, \emph{frozen}, in this study. Then, \proposal specializes the fully-connected part of the neural network of the target task by resorting to a GA. This metaheuristic wrapper prunes unnecessary neurons of these layers driven by the improvement of a performance measure (e.g. accuracy). The outcome of the process is a pruned network with a potentially improved accuracy by virtue of an \begin{idea2}evolved\end{idea2} pruning mask.
\end{idea}

\begin{figure}[H]
    \centering
    \subfloat[\centering Neuron Encoding]{ {\includegraphics[width=2.8cm]{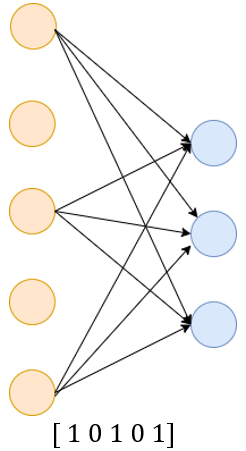}} }
    \qquad
    \subfloat[\centering Single Connection Encoding]{{\includegraphics[width=3.4cm]{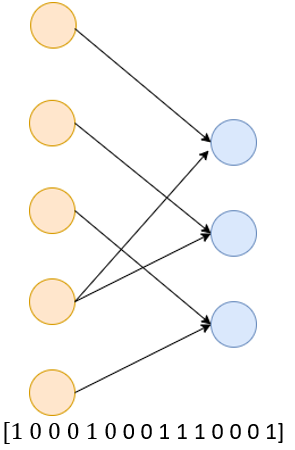}} \label{fig:encondingneurons}}
    \caption{Representation of encoding strategies}
    \label{fig:examples}
\end{figure}

In this study, we propose a novel method to prune the neurons, that considers the removal of both single connections and groups of connections of the input connections of a specific neuron, as can be observed in Fig. \ref{fig:examples}. \begin{idea}Fig \ref{fig:encondingneurons}\end{idea} shows a sparse layer that leads to the encoding strategy used in this work. This encoding, which is represented by the chromosome of the GA, is required to know exactly which connections are removed.

\proposal model utilizes a GA designed to \begin{idea2}evolve\end{idea2} the connections of a sparse layer. The GA takes each individual as a mask for the neural network and creates a sparse layer activating from the mask. This \begin{idea2}evolved\end{idea2} mask gives rise to a pruned neural network suitable for the problem under consideration.

The \begin{idea2}evolution of the connections\end{idea2} is performed using both methods, either by groups of connections or by single connections. The genome representation of each chromosome of the GA is binary-coded and represents the active \textit{neurons} or the active \textit{connections}. The GA evolves the configuration of the network towards its best pruned variant in terms of accuracy. Next, we describe both encoding strategies:

\begin{itemize}
\item \textbf{Neurons}: each gene of the chromosome represents the number of active neurons. A value 1 in position $i$ means that the neuron $i$ is active, and a 0 that is inactive. A non-active neuron implies that all the input connections are removed both in training and inference times. The length of the chromosome in this case is the number of neurons of the sparse model.

\item \textbf{Connections}: each gene represents the connection between the layers. The interpretation of the binary values is as follows: if a gene is 1, the connection between the corresponding layers exists, otherwise, that connection does not exist. \begin{idea}
Therefore, the length of the chromosome is the maximum number of connections, noted as $D = D_1 \times D_2$, where $D_1$ is the number of neurons in the previous layer, and $D_2$ is the number of neurons in that layer.
\end{idea}
\end{itemize}

An example of both encoding strategies is shown in Fig. \ref{fig:examples}. In both cases, the pruned connections are from the input on the layer, i.e. the right layer. The left image shows a representation of neuron-wise encoding, in which a group of neurons is selected to be active and the rest are pruned. The right image depicts how single connections are pruned.

\subsection{Evolutionary components of \proposal}\label{sec:ga-elements}

In this subsection, we introduce the evolutionary components of \proposal. It is a steady-state genetic algorithm, \begin{idea}which means that two new individuals, called offsprings, are created in each generation, \end{idea} for the previous mentioned encoding strategies (neuron encoding vs single connection encoding, Fig \ref{fig:examples}): in each iteration two individuals are selected and crossed, producing two offsprings that could \begin{idea}also\end{idea} be mutated. The offspring candidates are introduced in the population only if they improve the worst candidates in the population, replacing them.

As previously mentioned, in \proposal each chromosome is a binary array and each gene represents a connection between two layers. Each generation follows the classical scheme of selection, crossover, mutation and replacement. The best solutions found during the evolutionary search are kept in a population of individuals. Next, we describe the different components:

\textit{Selection}: the implemented selection operator is Negative Assorting Mating \begin{idea}(NAM)\end{idea} \citep{fernandes2001study}. The first parent is picked uniformly at random, while the second parent is selected between three possible candidates. \begin{idea} These candidates are also picked uniformly at random from the population. \end{idea} The candidate with higher Hamming distance from the first parent is chosen as the second parent, thereby ensuring that the recombined parents are diverse. This selection method allows for a higher degree of exploration of the search space. 

\textit{Crossover}: \proposal uses the uniform crossover operator shown in Expression \ref{crossover_operator}. Given two parents \textbf{P} and \textbf{Q}, where $\textbf{P}=\{p_i\}_{i=1}^D$ and $\textbf{Q}=\{q_i\}_{i=1}^D$. Then two offsprings $\textbf{P'}=\{p'_i\}_{i=1}^D$ and $\textbf{Q'}=\{q'_i\}_{i=1}^D$ are created following the equations:
\begin{equation}\label{crossover_operator}
  \begin{split}
        pi' = \left\{
            \begin{array}{ll}
            p_i & \mbox{if } r \le 0.5\\
            q_i & \mbox{otherwise}\\
            \end{array}
        \right. \\
        q_i' = \left\{
            \begin{array}{ll}
            q_i & \mbox{if } r \le 0.5\\
            p_i & \mbox{otherwise}\\
            \end{array}
        \right. 
  \end{split}
\end{equation}
where $r$ is the realization of a continuous random variable with support over the range $[0.0, 1.0]$. \begin{idea} This operator takes two parents $\textbf{P}$ and $\textbf{Q}$ of length $D$ and creates two new offspring $\textbf{P'}$ and $\textbf{Q'}$ of the same size. Each new offspring is composed of the parents genes. Each gene (position of the new array) is set equal to the gene of the first or second parent. This process is repeated until the whole offspring is composed.  \end{idea}

%An example of this operator is shown in Fig. \ref{fig:crossover}.
%\begin{figure}[H]
%  \centering
%  \includegraphics[width=0.5\textwidth]{images/crossover.png}
%  \caption{Uniform crossover operator.}
%  \label{fig:crossover}
%\end{figure}

\textit{Mutation}: \proposal adopts the so-called single point mutation. A mutation probability for each individual is defined by $p_{mut}$. Then, a gene of that individual is uniformly randomly selected and its bit is flipped\begin{idea}, i.e., if the mutation is performed, then that neuron or connection changes its value, which implies that the connection or the neurons is activated or deactivated. In this operator, $p_{mut}$ is the value that establishes the probability that a mutation is performed. \end{idea}

\textit{Replacement Strategy}: at the end of every generation, the two offsprings resulting from the crossover and mutation operators compete against the worst two elements. As a result, the population is updated with the best two individuals among them, i.e. those whose fitness value is better. \begin{idea} \proposal maintains a pool of four individuals: two offsprings and the two worst individuals selected from the population. Then, the best two of them are in the new population. The criterion to select the best two is based on the fitness as the best of them are selected. In case of same values, the individuals with fewer active neurons/connections are those selected to be retained in the new population. \end{idea}

\textit{Initialization}: the genes composing the individuals are initialized to 0 or 1 as per the following probabilistic condition with a $p_{one}$ probability:
\begin{equation} \label{initoperator}
  I_i=\left\{
    \begin{array}{ll}
      1 &\mbox{ if } r \le p_{one}\\
      0 & \mbox{ otherwise }\\
    \end{array}
  \right.
\end{equation}
where $r$ is the realization of a uniform continuous random variable with support $[0.0,1.0]$. 

\textit{Evaluation of individuals}: the fitness value of every individual is given by the accuracy over a test dataset of the neural network pruned as per the decoded individual, and trained over the training dataset of the task at hand. \begin{idea} Each individual, named $p$, is decoded to yield a sparse neural network, which we hereafter refer to \textit{SparseNet$_p$}. Then this network is trained as previously commented over the train dataset, giving the \textit{TrainedSparseNet$_p$} network. Lastly, the test dataset is evaluated in this network, producing the fitness of the individual, which we call \textit{ChildFitness$_p$}. \end{idea}

%(more details in Algorithm \ref{alg:eval_config}):
%\begin{enumerate}
%\item The sparsely layer of the network is set following the values of the chromosome.
%\item The network is trained from the training dataset.
%\item The fitness is the accuracy of the test dataset on the trained network.
%\end{enumerate}

\begin{idea} Algorithm \ref{alg:proposal} shows the pseudocode of \proposal. First, we need to understand what \proposal requires to start its evolution process, and what results from this process. The input of \proposal is determined by: 

    \begin{itemize}
        \item Dataset and task to be modeled.   
        \item Configuration of the GA: parameters needed for the algorithm.
        \item Configuration of the network: parameters needed for the network.
        \item Feature extractor: a pre-trained neural network used for feature extraction and TL, e.g., ResNet-50 trained over Imagenet or any other available architecture alike.
    \end{itemize}

The algorithm starts by initializing the individuals of the population (line 1) using the previous operator and then evaluating them (line 2). The evolutionary process is performed in lines 3-15. Two parents are selected using the NAM operator (line 4) and then the two offsprings are generated using the crossover operator (line 5). If the mutation condition is met, then mutation is performed (lines 6-8). The child population is now evaluated (lines 9-14). The evaluation is held over three steps, in which each individual is decoded (variable SparseNet$_p$ in line 10), and then the network created with its configuration is trained (named TrainedSparseNet$_p$ in line 11) using the train dataset and then evaluated (called ChildFitness$_p$ in line 12) over the test dataset. Lastly, the replacement strategy is triggered (line 15). The stopping criterion is the evaluation of a maximum number of networks.

\end{idea}

\begin{algorithm}[!hbt] 
\SetAlgoLined
 \caption{\proposal }
\label{alg:proposal}

\begin{idea}
\SetKwInOut{Input}{Input}
\SetKwInOut{Output}{Output}
\Input{Dataset, configuration of the GA, configuration of the network and feature extractor}
\Output{\begin{idea2}Evolved\end{idea2} pruned network}

 Initialization of individuals of the population using the initialization operator\;
 Evaluation of the initial population (see lines 9-14)\;
 \While{evaluations $<$ max\_evals} {
  Parent selection using NAM operator\;
  Generate offsprings using crossover operator\;
  \If{rand() $<$ p\_{mut}}{
    Perform mutation using mutation operator\;
  }
  \For{each child $p$ in children population} {
    SparseNet$_p$ $\leftarrow$ Create sparse network using the decoded individual of the population\;
    TrainedSparseNet$_p$ $\leftarrow$ Train SparseNet$_p$ using \textit{train dataset}\;
    ChildFitness$_p$ $\leftarrow$ Accuracy of TrainedSparseNet$_p$ evaluated in \textit{test dataset}\; 
    evaluations+=1\;
  }
  Replacement Strategy: child population vs worst individuals of population\;
 }
\end{idea} 
\end{algorithm}

\subsection{\proposal Network} \label{sec:network}

\begin{figure}[H]
    \centering
    \subfloat[\centering Pruning one layer]{{\includegraphics[width=7.5cm]{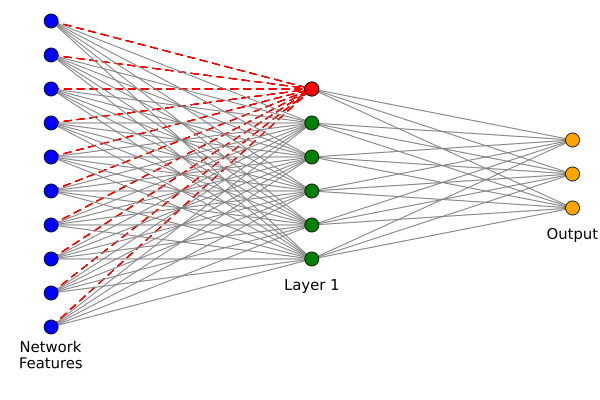}}
    \label{fig:pruningonefc}
    }
    \qquad
    \subfloat[\centering Pruning extracted features]{{\includegraphics[width=7cm]{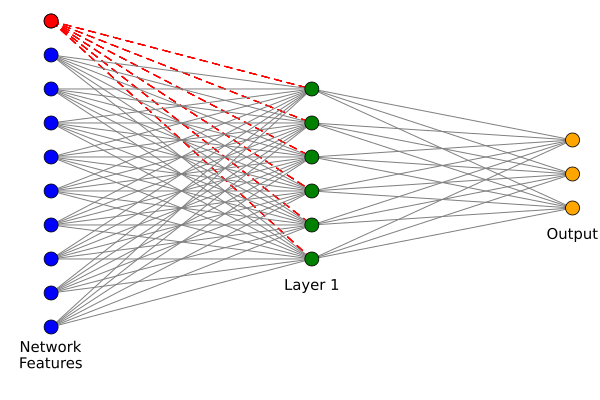}}
    \label{fig:pruningfs}
    }
    \caption{\begin{idea}Visualization of pruning architectures with one layer\end{idea}}
    \label{fig:onelayerarchitectures}
\end{figure}

\begin{figure}[H]
    \centering
    \subfloat[\centering Pruning the first layer]{{\includegraphics[width=7.5cm]{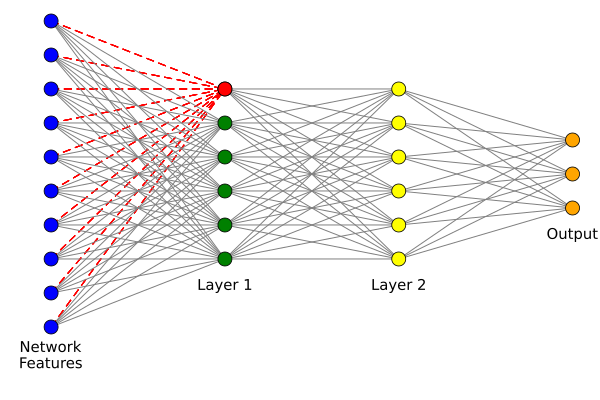}}
    \label{fig:pruningfirst}
    }
    \qquad
    \subfloat[\centering Pruning the second layer]{{\includegraphics[width=7.5cm]{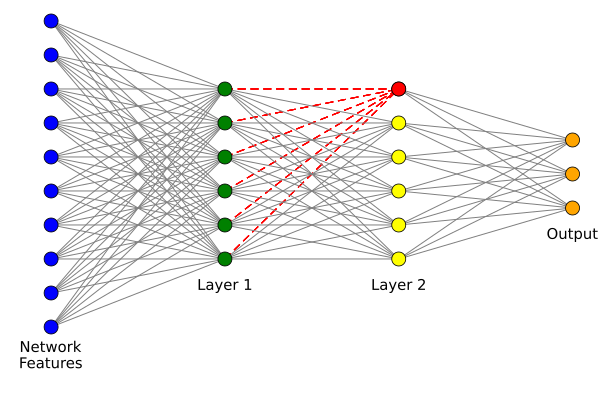}}
    \label{fig:pruningsecond}
    }
    \caption{\begin{idea}Visualization of pruning architectures with two layers\end{idea}}
    \label{fig:twolayerarchitectures}
\end{figure}

% primer parrafo -> uso de redes como extractores de características
\begin{idea}
This subsection is devised to fully understand the components associated with the networks that involve \proposal. \proposal stands for the usage of transfer learning, which means that the convolutional phase before the fully-connected layers is imported from other pre-trained model. Thus, the chosen CNN works as a feature extractor, i.e, obtains the main feature or characteristics for the task at hand. In our study, we have chosen ResNet-50 as feature extractor, although others can also be used, such as VGG or DenseNet. In Section \ref{sec:results}, a comparison between these three extractors is made to analyze the goodness of \proposal with them.

% segundo párrrafo -> composición de la red densa
These features are used as the input for the fully-connected layers. We introduce two different compositions of these fully-connected layers:

\begin{itemize}
    \item Single fully-connected layer: it is composed of a single layer with 512 neurons, followed by the output layer.
    \item Two fully-connected layers: this architecture has two layers of 512 each, and the output layer connecting the output of the last fully connected layer to as many neurons as the number of classes to be discriminated in the dataset.
\end{itemize} 

\begin{idea2}
    Moreover, activations of each fully connected layer are selected to be Rectifier Linear Unit (ReLU) functions. The output layer resorts to a SoftMax activation, which renders a probability distribution over the classes that compose the task at hand. The output neuron with the highest SoftMax probability leads to the predicted class of the input image.
\end{idea2}

% tercer párrafo -> importancia de la matriz de  adyacencia
\proposal stands for the usage of sparse layers. By definition, a sparse layer has few active connections. A key object in this environment is the adjacency matrix. This matrix is key in our study because it is used to create a sparse layer from it. It allows decoding an individual evolved via the GA to yield, as a result, a neural network with a sparse layer. Taking a look about this matrix, \proposal performs the pruning in relation to the connections/neurons that compose the input of the neuron. Consequently, there may be some neurons of the second layer which have no connection from the previous layer.

Based on the two network architectures described above, we consider different scenarios where \proposal can be applied. We present these scenarios in the following figures, in which the red-dashed lines indicate the effects of the pruning. In addition to that, we have grouped the models in terms of the number of the last layers. The first model is the application of the \proposal to prune model with one layer, which is shown in Fig. \ref{fig:pruningonefc}. Moreover, when the pruning is made with networks with two layers, three cases come up: pruning the first layer (see Fig. \ref{fig:pruningfirst}), pruning the second layer (see Fig. \ref{fig:pruningsecond} for these cases), or both at the same time, which is the combination of the last two cases. Lastly, \proposal is also able to prune the characteristics that are extracted from the network. This approximation is called Feature Selection because \proposal prunes the features that are less important to enhance the accuracy of the network. Fig. \ref{fig:pruningfs} illustrates how pruning in this last scenario reduces to a selection of features.
\end{idea}

\section{Experimental Framework} \label{sec:framework}

In this section, we describe the experimental framework followed in our study. First, we give a brief description of the analyzed datasets. Then, the training setup is presented, emphasizing the parameters of \proposal and the experimental conditions.

\subsection{Datasets}

In our study, we have chosen several diverse and representative datasets that are suitable for TL due to their size, as they require less training and inference time. Therefore, these datasets are suitable for population metaheuristics, as many individuals are evaluated. The selected data sets are shown in Table \ref{tbl:datasets}, which portrays their main characteristics for our experiments.

\begin{table}[!h]
\centering
\caption{Datasets used in the experiments.}
\label{tbl:datasets}
%\resizebox{\columnwidth}{!}{%
%    {
    \begin{tabular}{lccc}
    \toprule
         \textbf{Dataset} & \textbf{Image Size} & \textbf{\# classes} & \makecell{\textbf{\# Instances} \\ \textbf{(train / test)}} \\ \midrule
        SRSMAS        & $(299,299)$ & 14 & 333 / 76  \\
        RPS          & $(300,300)$ & 3 & 2520 / 372  \\
        LEAVES       & $(256,256)$ & 4 & 476 / 120 \\
        PAINTING     & $(256,256)$ & 5 & 7721 / 856 \\
        CATARACT      & $(256,256)$ & 4 & 480 / 121 \\
        PLANTS       & $(100,100)$ & 27 & 2340 / 236 \\
        \bottomrule
    \end{tabular}
%  } }
\end{table}

These datasets are diverse and taken from the literature:

\begin{itemize}
    \item SRSMAS \citep{SRSMAS} is a dataset to classify coral reef types with different classes and high distinction difficulty.
    \item RPS \citep{rps} is a dataset to identify the gesture of the hands in the popular Rock Paper Scissors game from images that have different positions and different skin colors.
    \item LEAVES is composed of images of healthy and unhealthy citrus leaves, with different shades of green \citep{RAUF2019104340}. 
    \item PLANTS is another dataset from the natural environment \citep{plantdoc}, in which the task is to differentiate between leaves of different plants such as tomato, apple or corn, among others.
    \item CATARACT comes from the medical domain \citep{Eyes}, whose purpose is to classify different types of eye diseases. 
    \item PAINTING is related to the painting world \citep{Musemart}. The images in this dataset have been taken from a museum and the task is to recognize different types of paintings.
\end{itemize}

\begin{figure}[!hbtp]
    \begin{center}
        \begin{minipage}[b]{.45\textwidth}
        \includegraphics[width=0.38\linewidth]{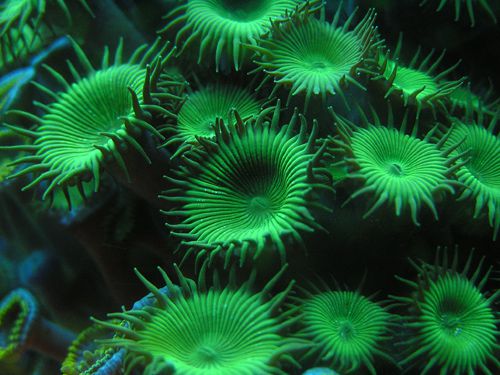}\qquad
        \includegraphics[width=0.38\linewidth]{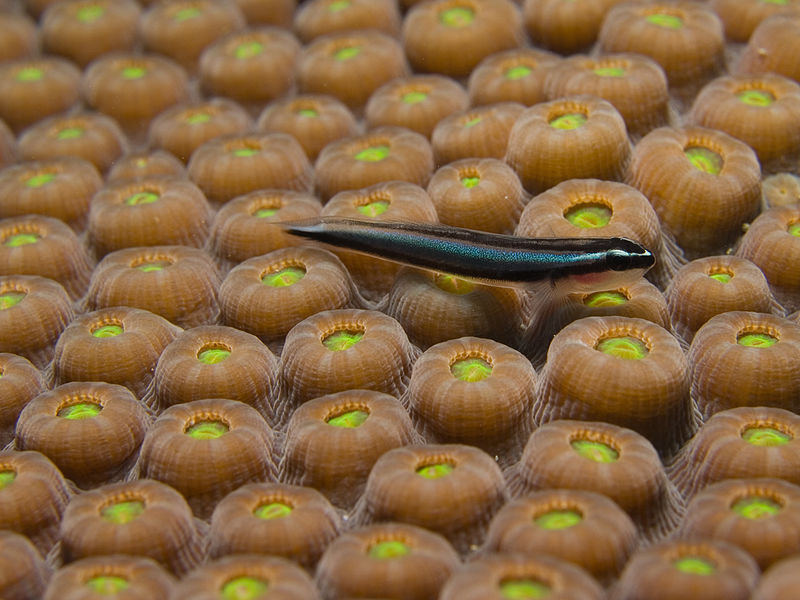}
        \end{minipage}

        \begin{minipage}[b]{.45\textwidth}
        \includegraphics[width=0.38\linewidth]{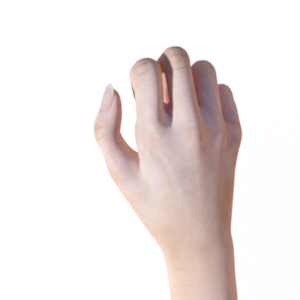} \qquad
        \includegraphics[width=0.38\linewidth]{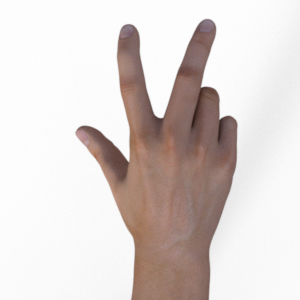}
        \end{minipage}
    
        \begin{minipage}[b]{.45\textwidth}
        \includegraphics[width=0.38\linewidth]{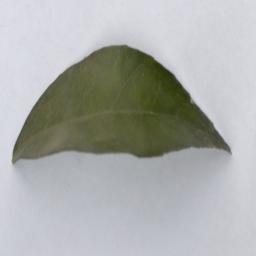}\qquad
        \includegraphics[width=0.38\linewidth]{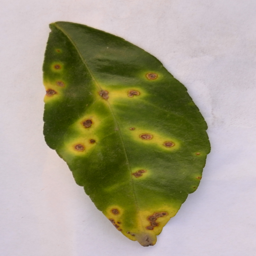}
        \end{minipage}
    \end{center}
\caption{Images of datasets. Top: SRSMAS examples. Middle: RPS examples. Bottom: LEAVES examples.}
\label{fig:images-datasets}
\end{figure}

Examples for several of the above datasets are shown in Fig. \ref{fig:images-datasets}.

\subsection{Training setup}\label{training}

The evaluation of \proposal requires splitting the images of the datasets in train and test subsets. As the results could strongly depend on the train and test sets, we have applied in SRSMAS and LEAVES a 5-fold cross-validation \footnote{Sets for 5 fold CV for SRSMAS and LEAVES: \\https://drive.google.com/drive/folders/1Xf7OeZyWDDG-\_Y4VX\_nnAdfz3Kwhy8LU?usp=sharing. Last Access: 28/01/2022}

For the remaining datasets, the train and test had already been defined beforehand, so we have used them for the sake of replicability.

The training is done using SGD as optimizer, a batch size of 32 images, and maximum 600 epochs, but the training stops when no improvement of loss is obtained in ten consecutive epochs. The model with the greater accuracy on the training set is saved. As we apply TL, only the last layers are trained, whereas the remaining ones are frozen with the parameter values imported from the pre-trained ResNet-50 network.

\begin{table}[!h]
  \centering 
  \caption{Parameters of \proposal.}
  \label{tbl:parameters}
  \begin{tabular}{cc}
    \toprule
    \textbf{Parameter} & \textbf{Value}\\
    \midrule
    \multirow{2}{*}{Maximum Evals} & 200 (one layer) \\
                                & 300 (both layer) \\
    \# Runs & 5\\
    Population size &  30\\
    NAM & 3\\
    $p_{mut}$ & 0.07\\
    Batch Size & 32 \\
    \bottomrule
  \end{tabular}
\end{table}

The parameters of \proposal are indicated in Table \ref{tbl:parameters}. We have set the maximum evaluations to two different values, 200 and 300, because there are some experiments we carry out to analyze the behavior of \proposal that need an adaptation of this value because the search space in these experiments is wider. \begin{idea} The size of the population of networks that our model evolves at each generation is set to 30, the mutation probability is $p_{mut}$ and the NAM operator chooses the second parent among 3 candidates. \end{idea} The best solution found in terms of accuracy is returned. We note that in case of several solutions with the same accuracy, the returned solution is the configuration with the lowest percentage of active neurons. Note that the number of runs and total function evaluations is kept low to meet a computationally affordable balance between performance and the high execution times required for simulation. This is shown in Table \ref{tbl:alltimeproposal}, in which the average time per execution of the models with two layers is indicated. Unfortunately, this limited number of runs per experiment impedes the application of statistical tests to assess the significance of the reported differences, as tests conventionally used for this purpose require larger sample sizes to reach meaningful conclusions.

\begin{table}[!h]
  \centering
  \caption{Average time per run of \proposal.}
  \label{tbl:alltimeproposal}
  %\resizebox{\columnwidth}{!}{%
  %{
  \begin{tabular}{crrr}
    \toprule
    Dataset & First Layer & Second Layer & Both Layers\\
    \midrule
    SRSMAS & 9h 41min & 10h 45min & 15h 27min \\
    RPS & 4h 23min & 5h 13min & 8h 00min \\
    LEAVES & 10h 26min & 16h 01min & 17h 45min \\
    PAINTING & 13h 1min & 15h 24min & 22h 45min \\
    CATARACT & 2h 3min & 2h 22min & 3h 26min \\
    PLANTS & 6h 3min & 6h 11min & 10h 00min \\
    \bottomrule
  \end{tabular}
  %}}
\end{table}

All the following experiments have been carried out using Python 3.6 and a Keras/Tensorflow implementation deployed and running on a Tesla V100-SXM2 GPU. The code is published in a open repository in GitHub.\footnote{\proposal repository: https://github.com/ari-dasci/S-EvoDeepTLPruning.}

\section{Results and Discussion} \label{sec:results}

In this section, we analyze the behavior of \proposal. In order to show the benefits of using \proposal, we propose four research questions (RQ) that they are going to be answered with different and diverse experiments over several datasets are carried out. We will show tables with the results of these experiments and we will analyze them to ensure the benefits of \proposal. These RQ are the following ones:

\begin{enumerate}[start=1,label={(\bfseries RQ\arabic*)}]
    \item Which is the performance of \proposal against fully-connected models? \newline \newline
    We compare \proposal against non-pruned models comprising fully-connected layers to study which model obtains a better performance in the experiments. Moreover, we remark the flexibility of \proposal applying it with one and several layers.
    
    \item Which would be better, to remove neurons or connections? \newline \newline
    We compare \proposal using the two alternatives explained in the previous section: 1) pruning the neurons or 2) each individual represents exact connections between neurons, allowing for a more finely grained \begin{idea2}evolution of the connections\end{idea2}. The goal of this section is to check which representation obtains the best results. On the one hand, the neuron representation of the length of chromosomes is shorter, so the domain search is smaller. On the other hand, the connections representation is a more fine-detail representation, so it could potentially allow the algorithm to obtain better results.

    \item Which is the performance of \proposal when compared to efficient pruning methods? \newline \newline We compare the performance of \proposal against several efficient pruning methods published in the literature for compressing CNN networks: Polynomial Decay \citep{zhu2017prune}, Weight Pruning \citep{han2015learning} and Neuron Pruning \citep{srinivas2015data}. This comparison of \proposal and the CNN models is made in terms of accuracy and model compression.
    
    \item Which would be better, pruning fully-connected layers or performing Feature Selection? \newline \newline A particular case of \proposal stands when the optimization of the network is done with one fully-connected layer and features are evolved towards the fittest for the problem at hand. Our aim is to check if this scheme improves the overall performance of the model over a dataset in terms of the accuracy of the models.
    
    \begin{idea}
    \item How does \proposal perform when applied to different pre-trained networks? \newline \newline We compare \proposal with different feature extractors. From RQ1 to RQ4, ResNet-50 is used to extract the features or characteristics for the considered datasets. Experiments devised for this RQ aim to examine whether \proposal adapts suitably to other feature extractors such as DenseNet-121 and VGG-19, so that better-performing pruned networks are produced by our proposal also for these feature extractors.
    
    \item Can \proposal adapt efficiently their pruned knowledge to changes in the modeling task, showing robustness? \newline \newline We analyze the behavior of \proposal when the datasets change. In this case, we have selected some datasets from our study, and we have done several modifications, removing partially or totally a class. Within these changes, we want to show both the robustness of the \proposal in different situations and that the pruning models obtained by the GA of \proposal have been adapted to each one of these situations.
    \end{idea}
\end{enumerate}

This section is divided in Section \ref{sec:rd-pruning}, where the comparison of the diverse representations of pruning that \proposal makes against the reference models is presented to answer RQ1. Next, Section \ref{sec:rd-neuronsorconnections} discusses whether \proposal should operate over neurons or connections to analyze RQ2. Section \ref{sec:rd-comparison} provides a complete comparison among \proposal and other efficient pruning methods in order to solve RQ3. Section \ref{sec:rd-fs} explains the approximation of Feature Selection. A whole comparison against all the previous models is made to assess the importance of the Feature Selection to answer RQ4. \begin{idea} Section \ref{sec:rd-extractors} shows the comparison of the best two models of \proposal with different feature extractors. Lastly, in Section \ref{sec:rd-generalization}, \proposal is challenged, with several modifications of the used datasets, to grasp the relevant features of these datasets and to analyze the robustness of our proposal.\end{idea}

\subsection{Answering RQ1: Pruning} \label{sec:rd-pruning} 

In this section, we assess the performance gaps between the proposed \proposal against other reference models to answer  RQ1. In each subsection, several and diverse experiments are carried out to present results that assure the quality of \proposal when it is compared to other models. This pruning section is composed of Section \ref{sec:proposal1Layer}, in which we compare \proposal against reference models with only one layer; of Section \ref{sec:proposal2Layers} we make the same experiments but with two layers and the \begin{idea2}evolution\end{idea2} of one of them, and of Section \ref{sec:proposalboth}, that shows the \begin{idea2}evolution\end{idea2} of two consecutive layers at the same time.

In the following, we describe the different reference models:

\begin{itemize}
    \item The first reference is composed of fully-connected layers of 512 units and the output layer. That is equivalent to the model with all neurons in active mode (all gens to 1). This model is the one with all active neurons, we call it \textit{Not Pruned}.
    
    %\item The second one is the previous model with the addition of a standard dropout. We call this model \textit{Dropout}.
    \item A grid search scheme is compared to \proposal to check whether the improvement made by \proposal could be obtained with a simple search over the percentage of neurons of the fully-connected layer. We have tested the fully-connected model with different number of neurons: 10\% to 90\% of its total units increasing this percentage by 10\% (including both), and for each dataset we have identified the number of neurons which gives the best accuracy. 

    %\item The fourth full-connected model has as many neurons as the best model in terms of accuracy is obtained by \proposal. This model is called \textit{Dense Neurons} 
    \item The best result of the above models is also noted and it is called \textit{Best Fixed}. When implemented over both layers, pruning is referred to as \textit{Best Fixed Both}.
\end{itemize}

\subsubsection{Pruning neurons of one fully-connected layer} \label{sec:proposal1Layer}

This section introduces the results of pruning models with only one fully-connected layer. Table \ref{tab:proposal-one-layer} shows the comparison of \proposal against the reference models. In this case, the reference model only has one fully-connected layer composed by 512 units and the output layer. 

For each dataset, the first row shows the obtained average accuracy by the models over the test set, whereas the second row informs about the average percentage of active neurons.

\begin{table}[!h]
\centering
\caption{Average results of \proposal against not pruned models with one fully-connected layer.}
\label{tab:proposal-one-layer}
%\resizebox{\columnwidth}{!}{%
%{
\begin{tabular}{*{1}{c}*{4}{c}}
  \toprule
  % Cambiar orden del nombre
    \multirow{2}{*}{Dataset} & \multirow{2}{*}{Measure} & \multirow{1}{*}{Not} &   \multirow{1}{*}{Best} & \multirow{1}{*}{EvoPrune} \\
    &  & \multirow{1}{*}{Pruned} & \multirow{1}{*}{Fixed}& \multirow{1}{*}{DeepTL}\\
    \midrule
    \multirow{2}{*}{SRSMAS} & Accuracy & 0.832 &  0.866 &  \textbf{0.885} \\
                          & \% Active neur. & 100  & 20 & 25 \\
    \midrule
    \multirow{2}{*}{RPS} & Accuracy & 0.938 & 0.938 & \textbf{0.954} \\
                       & \% Active neur. & 100 & 40 & 46 \\
    \midrule
    \multirow{2}{*}{LEAVES} & Accuracy & 0.923 & 0.927 &  \textbf{0.935} \\
                          & \% Active neur. & 100 & 80 & 38  \\
    \midrule
    \multirow{2}{*}{PAINTING} & Accuracy& 0.939 & 0.945 &  \textbf{0.951}  \\
                              & \% Active neur. & 100 & 60  & 46  \\
    \midrule    
    \multirow{2}{*}{CATARACT} & Accuracy & 0.703 & 0.719 & \textbf{0.732}  \\
                          & \% Active neur. & 100  & 70 & 39  \\
    \midrule    
    \multirow{2}{*}{PLANTS} & Accuracy & 0.432 & 0.432 &  \textbf{0.480} \\
                          & \% Active neur. & 100 & 10 &  49  \\
    \midrule
\end{tabular}
%}}
\end{table}

These results show how \proposal is capable of distinguishing the pruning configurations that lead towards an improvement of performance of the models, as it obtains a greater accuracy in all the datasets for every reference model. Moreover, in most datasets\begin{idea},\end{idea} a higher compression ratio than the best fully-connected model is also achieved.

\subsubsection{Pruning neurons of two fully-connected layers} \label{sec:proposal2Layers}

In this section, our challenge is to improve the performance of a two fully-connected layer network. For that reason, \proposal is applied to each layer individually. 

The results of applying \proposal to each layer individually are shown in Table \ref{tab:proposal-two-layers}, where \textbf{First Layer} indicates the case of the \begin{idea2}evolution\end{idea2} of the first layer, and \textbf{Second Layer} describes the other case. In this case, Both \textit{Not Pruned} and \textit{Best Fixed} are the reference models with two fully-connected layer.

In this case, results follow the same path as the previous one: in all the datasets\begin{idea},\end{idea} \proposal achieves an improvement of the accuracy over the reference models. Moreover, the Second Layer case obtains more compressed networks than the First Layer option.

Comparing the results of the scheme of one and two layers, both have similar results, only in RPS and CATARACT the difference in terms of accuracy is higher. Thus, these experiments have shown the ability of \proposal of improving the overall performance of networks and, at the same time, reducing their complexity.

\begin{table*}[!htp]
\centering
\caption{Average results of \proposal against not pruned models with two fully-connected layers.}
\label{tab:proposal-two-layers}
%\resizebox{\textwidth}{!}{%
%{
\begin{tabular}{llcccccc}
\toprule
& & \multicolumn{3}{c}{\textbf{First Layer}} & \multicolumn{3}{c}{\textbf{Second Layer}} \\
\cmidrule(lr){3-5}\cmidrule(lr){6-8}

\multirow{2}{*}{Dataset} & \multirow{2}{*}{Measure} & Not & Best & EvoPrune & Not & Best & EvoPrune \\
& & Pruned & Fixed & DeepTL & Pruned & Fixed & DeepTL \\

\cmidrule(lr){1-5}\cmidrule(lr){6-8}

\multirow{2}{*}{SRSMAS} & Accuracy & 0.858 & 0.858 & \textbf{0.883} & 0.858 & 0.860 & \textbf{0.884}  \\
   & \% Active neur. & 100 & 100 & 46 & 100 & 80 & 47 \\

\cmidrule(lr){1-5}\cmidrule(lr){6-8}

\multirow{2}{*}{RPS} & Accuracy & 0.922 & 0.938 & \textbf{0.959} & 0.922 & 0.949 & \textbf{0.969} \\
& \% Active neur. & 100 & 30 & 37 & 100 & 30 & 16 \\

\cmidrule(lr){1-5}\cmidrule(lr){6-8}

\multirow{2}{*}{LEAVES} &  Accuracy & 0.919 & 0.926 & \textbf{0.937} & 0.919 & 0.929 & \textbf{0.935} \\
   & \% Active neur. & 100 & 40 & 28 & 100 & 60 & 12 \\

\cmidrule(lr){1-5}\cmidrule(lr){6-8}

\multirow{2}{*}{PAINTING} &  Accuracy & 0.939 & 0.944 & \textbf{0.950} & 0.939 & 0.941 & \textbf{0.951} \\
   & \% Active neur. & 100 & 60 & 53 & 100 & 90 & 53 \\

\cmidrule(lr){1-5}\cmidrule(lr){6-8}

\multirow{2}{*}{CATARACT} & Accuracy & 0.703 & 0.711 & \textbf{0.740} & 0.703 & 0.703 & \textbf{0.735} \\
   & \% Active neur. & 100 & 70 & 63 & 100 & 100 & 59 \\

\cmidrule(lr){1-5}\cmidrule(lr){6-8}

\multirow{2}{*}{PLANTS} & Accuracy & 0.402 & 0.448 & \textbf{0.479} & 0.402 & 0.441 & \textbf{0.483} \\
   & \% Active neur. & 100 & 10 & 45 & 100 & 50 & 37 \\
\bottomrule
\end{tabular}
%}}
\end{table*}

\subsubsection{Pruning neurons of both layers} \label{sec:proposalboth}

In the previous sections\begin{idea},\end{idea} we have tested \proposal to \begin{idea2}solve problems via the evolution of a single layer\end{idea2}. In this section we increase the difficulty of the problem: the \begin{idea2}evolution\end{idea2} of two consecutive fully-connected layers.

From the previous experiments, we have run \proposal with 200 evaluations, but we have noticed that this number of evaluations might not be enough. This is due to the fact that we have now individuals of size 1024, 512 for each layer, and the search space is larger than in the rest of experiments. We have therefore also carried out the experiments with 300 function evaluations.

In Table \ref{tab:comparison-reference-consecutive}, we show the results for reference models and \proposal with 300 evaluations. The reference models stand the same as in the previous cases, but as they are implemented over both layers, the pruning is now referred to as \textit{Best Fixed Both}.

In some cases, the percentage of remaining active neurons is higher than in the first and second layer models, but that is due to the complexity of this new problem. However, the performance of the network in these experiments indicates that the best option for pruning is achieved when the \begin{idea2}evolution\end{idea2} is done to two consecutive layers.

\begin{table}[H]
\centering
\caption{Average results of \proposal against not pruned methods \begin{idea2}evolving\end{idea2} two consecutive layers.}
\label{tab:comparison-reference-consecutive}
%\resizebox{\columnwidth}{!}{%
%{
\begin{tabular}{*{1}{c}*{4}{c}}
  \toprule
  % Cambiar orden del nombre
    \multirow{2}{*}{Dataset} & \multirow{2}{*}{Measure} & \multirow{1}{*}{Not} &   \multirow{1}{*}{Best Fixed} & \multirow{1}{*}{EvoPrune} \\
    &  & \multirow{1}{*}{Pruned} & \multirow{1}{*}{Both}& \multirow{1}{*}{DeepTL}\\
  \midrule
  \multirow{2}{*}{SRSMAS} & Accuracy        & 0.858 &  0.863 &  \textbf{0.885} \\
                          & \% Active neur. & 100   & 50 & 64 \\
  \midrule
  \multirow{2}{*}{RPS} & Accuracy & 0.922 &  0.946 & \textbf{0.978} \\
                       & \% Active neur. & 100 & 90 & 12 \\
  \midrule
  \multirow{2}{*}{LEAVES} & Accuracy & 0.919 & 0.934 &  \textbf{0.936}  \\
                          & \% Active neur. & 100  & 15 & 34  \\
  \midrule
     \multirow{2}{*}{PAINTING} & Accuracy & 0.939 & 0.949 & \textbf{0.953} \\
                                & \% Active neur. & 100  & 40 & 51  \\
    \midrule    
    \multirow{2}{*}{CATARACT} & Accuracy & 0.703 & 0.735 & \textbf{0.746}  \\
                          & \% Active neur. & 100 & 85 & 63  \\
    \midrule    
    \multirow{2}{*}{PLANTS} & Accuracy & 0.402 & 0.466 & \textbf{0.491}  \\
                          & \% Active neur. & 100  & 55 & 41  \\
    \midrule    
\end{tabular}
%}}
\end{table}

These results prove the attainment of a sequential process to make pruning of DL models by adding layers and then, evolving their neurons to achieve a reduced configuration of the network. This process rises the performance of the models in terms of accuracy.

\subsection{Answering RQ2: which would be better, to remove neurons or connections? } \label{sec:rd-neuronsorconnections}

This section is devised to formally answer the RQ2, which is to decide if it is better \begin{idea} to \end{idea}perform pruning of whole neurons or either single connections, by comparing different \proposal chromosome representations: neurons and connections, as we described in Section \ref{sec:proposal}. Two representations are shown in this section: the neuron representation, in which a gen represents the connections of a neuron, and the connections representation, in which a gene represents a specific connection in the sparse layer. Neuron representation obtains shorter chromosomes than the connections one. Meanwhile, the connection representation leads to a more detailed representation and a larger domain search.

\begin{table*}[!h]
\centering
\caption{Average results of \proposal against edges models.}
\label{tab:proposal-vs-edges}
\resizebox{\textwidth}{!}{%
{
\begin{tabular}{llccccc}
\toprule
& & \multicolumn{2}{c}{\textbf{One Layer}} & \multicolumn{3}{c}{\textbf{Two Layers}} \\
\cmidrule(lr){3-4}\cmidrule(lr){5-7}

\multirow{2}{*}{Dataset} & \multirow{2}{*}{Measure} & \multirow{2}{*}{Edges} & \multirow{2}{*}{\proposal} & \multirow{2}{*}{Edges} & \proposal & \proposal\\
& & &  & & Layer 1 & Layer 2 \\

\cmidrule(lr){1-4}\cmidrule(lr){5-7}

\multirow{2}{*}{SRSMAS} & Accuracy & 0.875 & \textbf{0.885} & 0.875 & 0.883 & \textbf{0.884} \\
   & \% Active neur. & 43 & 25 & 46 & 46 & 47 \\

\cmidrule(lr){1-4}\cmidrule(lr){5-7}

\multirow{2}{*}{RPS} & Accuracy & 0.952 & \textbf{0.954} & 0.952 & 0.959 & \textbf{0.969} \\
& \% Active neur. & 29 & 46 & 37 & 37 & 16 \\

\cmidrule(lr){1-4}\cmidrule(lr){5-7}

\multirow{2}{*}{LEAVES} &  Accuracy & 0.932 & \textbf{0.935} & 0.933 & \textbf{0.937} & 0.935 \\
   & \% Active neur. & 45 & 38 & 45 & 28 & 12 \\

\cmidrule(lr){1-4}\cmidrule(lr){5-7}

\multirow{2}{*}{PAINTING} &  Accuracy & 0.949 & \textbf{0.951} & 0.950 & 0.950 & \textbf{0.951} \\
   & \% Active neur. & 48 & 46 & 53 & 48 & 53  \\

\cmidrule(lr){1-4}\cmidrule(lr){5-7}

\multirow{2}{*}{CATARACT} & Accuracy & 0.729 & \textbf{0.732} & 0.737 & \textbf{0.740} & 0.735 \\
   & \% Active neur. & 69 & 49 & 66 & 63 & 59 \\

\cmidrule(lr){1-4}\cmidrule(lr){5-7}

\multirow{2}{*}{PLANTS} & Accuracy & 0.457 & \textbf{0.480} & 0.463 & 0.479 & \textbf{0.483} \\
   & \% Active neur. & 64 & 49 & 45 & 45 & 37 \\
\bottomrule
\end{tabular}
}}
\end{table*}

Table \ref{tab:proposal-vs-edges} shows for each dataset and representation the mean accuracy and \% of active connections for both pruning methods. The connection strategy is named \textit{Edges}. The results show that even though there are some cases in which the edges \begin{idea2}evolution\end{idea2} achieves a similar performance of the network, the \begin{idea2}evolution of the neurons\end{idea2} presents more robust results. The models working at the neuron level are even able to further reduce the number of active neurons in some datasets.

As a conclusion of this experiment, we can confirm that using the neuron approach is the best representation and that the second layer model gives us more consistent results than the first layer pruning model, both in accuracy and in reduction of the model.

\subsection{Answering RQ3: Comparing \proposal with efficient methods for CNN pruning} \label{sec:rd-comparison}

This section is devised to analyze the RQ3 comparing \proposal to other well known network pruning methods to present results that measure the performance of our model against these methods. This comparison is conducted in terms of quality and computational complexity, aimed to prove the potential of \proposal with respect to other pruning counterparts. To this end, we implement two different pruning methods, namely, weight pruning and neuron pruning. These methods have a parameter in common, $S_f \in \mathbb{R}(0, 1)$, which denotes the target pruning percentage. It is set to the same percentage of reduction that \proposal has obtained in the experiments discussed previously. Next, we briefly describe each of such methods:

\begin{itemize}
    \item \texttt{weight} \citep{han2015learning}: Parameters with lower values are pruned at once. This method operates over the whole parameter set in the layer to be optimized. (Parameters: $S_f$)
    \item \texttt{polynomial decay} \citep{zhu2017prune}: Parameters are pruned guided by a Polynomial Decay schedule to the specified sparsity value. Between pruning steps, the network is allowed to fine tune for 5 epochs. This model is also applied over the whole parameter set in the layer to be optimized. Parameters used in the experimentation are listed in Table \ref{tab:poly-decay-params}.
    \item \texttt{neuron} \citep{srinivas2015data}: Neurons with lower mean input connection values are pruned. (Parameters: $S_f$) as in Figs. \ref{fig:onelayerarchitectures} and \ref{fig:twolayerarchitectures}. (Parameters: $S_f$)
\end{itemize}

Table \ref{tab:poly-decay-params} summarizes the value of the parameters of Polynomial Decay algorithm, which have been adapted to our experiments. Then, given a desired sparsity value of $S$, the sparsity is updated over a span of $k$ pruning steps following the next equation 
\small\begin{equation} \label{eq:poly}
    S_k = S_f + (S_i - S_f) \cdot \Big(1 - \frac{K_k - K_i}{K_f - K_i}\Big)^{\alpha} \text{ if } K_k \bmod F= 0 
\end{equation}\normalsize 
wherein parameters are described as follows:
\begin{itemize}
    \item $S_{i,f} \in \mathbb{R}(0, 1)$ are the initial and final sparsity percentages.
    \item $S_f$ depends on the experiment. It is the percentage of pruning that \proposal has achieved and the end of the generations.
    \item $K_{i,f} \in \mathbb{N}$ configures at what training step the pruning algorithm starts and ends.
    \item $K_k \in \mathbb{N}(K_i, K_f)$ is the current step.
    \item $nb$ is the number of batches. It is calculated as the length of the training set divided by the batch size.
    \item $F$ configures the frequency at which Equation \ref{eq:poly} is computed.
\end{itemize}

% esta es la tabla X
\begin{table}[!h]
  \centering
  \caption{Parameter values of Polynomial Decay.}
  \label{tab:poly-decay-params}
  \begin{tabular}{cc}
    \toprule
    \textbf{Parameter} & \textbf{Value}\\
    \midrule
    $S_i$ & 0.1\\
    $K_i$ & 0 \\
    $K_f$ & $nb\cdot25$ \\
    $F$ & $nb\cdot5$ \\
    $\alpha$ & 3.0 \\
    \bottomrule
  \end{tabular}
\end{table}

Parameters of the Polynomial Decay model are chosen to achieve a tradeoff between network recovery and the number of training epochs. Given the nature of this model, Polynomial Decay implies more training epochs than the implemented neuron and weight pruning methods. This fact could make the comparison between such methods unfair if the additional training epochs introduced by the Polynomial Decay model are high compared to the initial training epochs (i.e. 600). To avoid this situation, Polynomial Decay is configured so that it sufficiently guarantees network recovery for all datasets while a minimal amount of extra training epochs are carried out, just an extra 4\% from the initial 600 epochs (i.e. 25 extra epochs).

Our analysis aims to verify whether the performance of the above efficient pruning methods are comparable to \proposal in terms of solution quality (accuracy) when they are configured to prune the same amount of parameters. Thus, the experimentation is carried out for the previously four cases discussed, selecting the average outcomes from the experimentation conducted in this point. 

First, we show the results of this comparison when only a fully-connected layer is \begin{idea2}evolved\end{idea2}. Table \ref{tab:mean-cnn-evodeeptlpruning-1layer} shows the results for this case. \proposal outperforms the CNN models in five out of the six cases, but only in PAINTING these results are better for the Polynomial Decay or Weight models.

\begin{table}[!h]
\centering
\caption{Average results of \proposal against efficient CNN for one layer models.}
\label{tab:mean-cnn-evodeeptlpruning-1layer}
%\resizebox{\columnwidth}{!}{%
%{
\begin{tabular}{llcccc}
\toprule
& &  \multicolumn{4}{c}{\textbf{One Layer}} \\
\cmidrule(lr){3-6}

\multirow{2}{*}{Dataset} & \multirow{2}{*}{Measure} & \multirow{2}{*}{Weight} & Poly. & \multirow{2}{*}{Neuron} & EvoPrune\\

& & &  Decay &  & DeepTL \\

\cmidrule(lr){1-6}

\multirow{2}{*}{SRSMAS} & Accuracy & 0.805 & 0.823 & 0.745 & \textbf{0.885}  \\
   & \% Active neur. & 25 & 25 & 25 & 25 \\

\cmidrule(lr){1-6}

\multirow{2}{*}{RPS} & Accuracy & 0.917 & 0.927 & 0.869 & \textbf{0.954}  \\
& \% Active neur. &  46 & 46 & 46 & 46 \\

\cmidrule(lr){1-6}

\multirow{2}{*}{LEAVES} &  Accuracy & 0.918 & 0.920 & 0.886 & \textbf{0.935}  \\
   & \% Active neur. & 38 & 38 & 38 & 38 \\

\cmidrule(lr){1-6}

\multirow{2}{*}{PAINTING} &  Accuracy & 0.993 & \textbf{0.994} & 0.874 & 0.951  \\
   & \% Active neur. & 46 & 46 & 46 & 46 \\

\cmidrule(lr){1-6}

\multirow{2}{*}{CATARACT} & Accuracy & 0.678 & 0.679 & 0.658 & \textbf{0.732}  \\
   & \% Active neur. &  39 & 39 & 39 & 39 \\

\cmidrule(lr){1-6}

\multirow{2}{*}{PLANTS} & Accuracy & 0.406 & 0.411 & 0.365 & \textbf{0.480}  \\
   & \% Active neur. & 49 & 49 & 49 & 49 \\
\bottomrule
\end{tabular}
%}}
\end{table}

%% Tabla de EvoDeepTLPruning vs modelos de pruning con 2 CAPAS.
\begin{table*}[ht!]
\centering
\caption{Average results of \proposal against efficient CNN pruning methods for two layers models.}
\label{tab:mean-cnn-evodeeptlpruning}
\resizebox{\textwidth}{!}{%
{\begin{tabular}{llccccccccccccc}
\toprule
& & \multicolumn{4}{c}{\textbf{First Layer}} & \multicolumn{4}{c}{\textbf{Second Layer}} &\multicolumn{4}{c}{\textbf{Both Layers}}\\
\cmidrule(lr){3-6}\cmidrule(lr){7-10}\cmidrule(lr){11-14}

\multirow{2}{*}{Dataset} & \multirow{2}{*}{Measure} & \multirow{2}{*}{Weight} & Poly. & \multirow{2}{*}{Neuron} & EvoPrune & \multirow{2}{*}{Weight} & Poly. & \multirow{2}{*}{Neuron} & EvoPrune & \multirow{2}{*}{Weight} & Poly. & \multirow{2}{*}{Neuron} & EvoPrune \\
& & & Decay & & DeepTL & & Decay & & DeepTL & & Decay &  & DeepTL \\

\cmidrule(lr){1-6}\cmidrule(lr){7-10}\cmidrule(lr){11-14}

\multirow{2}{*}{SRSMAS} & Accuracy & 0.795 & 0.815 & 0.775 & \textbf{0.883} & 0.834 & 0.837 & 0.779 & \textbf{0.884} & 0.845 & 0.847 & 0.647 & \textbf{0.885}  \\
   & \% Active neur. & 46 & 46 & 46 & 46 & 47 & 47 & 47 & 47 & 64 & 64 & 64 & 64\\

\cmidrule(lr){1-6}\cmidrule(lr){7-10}\cmidrule(lr){11-14}

\multirow{2}{*}{RPS} & Accuracy & 0.886 & 0.911 & 0.803 & \textbf{0.959} & 0.845 & 0.911 & 0.696 & \textbf{0.969} & 0.694 & 0.899 & 0.490 & \textbf{0.978}\\
& \% Active neur. & 37 & 37 & 37 & 37 & 16 & 16 & 16 & 16 & 12 & 12 & 12 & 12\\

\cmidrule(lr){1-6}\cmidrule(lr){7-10}\cmidrule(lr){11-14}

\multirow{2}{*}{LEAVES} & Accuracy & 0.913 & 0.918 & 0.812 & \textbf{0.937} & 0.904 & 0.919 & 0.712 & \textbf{0.935} & 0.911 & 0.925 & 0.747 & \textbf{0.936}  \\
& \% Active neur. & 28 & 28 & 28 & 28 & 12 & 12 & 12 & 12 & 34 & 34 & 34 & 34  \\

\cmidrule(lr){1-6}\cmidrule(lr){7-10}\cmidrule(lr){11-14}

\multirow{2}{*}{PAINTING} & Accuracy & \textbf{0.995} & 0.993 & 0.850 & 0.950 & 0.937 & 0.938 & 0.920 & \textbf{0.951} & 0.934 & 0.940 & 0.853 & \textbf{0.953}   \\
    & \% Active neur. & 53 & 53 & 53 & 53 & 53 & 53 & 53 & 53 & 51 & 51 & 51 & 51   \\

\cmidrule(lr){1-6}\cmidrule(lr){7-10}\cmidrule(lr){11-14}

\multirow{2}{*}{CATARACT} & Accuracy & 0.668 & 0.684 & 0.673 & \textbf{0.740} & 0.694 & 0.689 & 0.648 & \textbf{0.737} & 0.686 & 0.696 & 0.611 & \textbf{0.746} \\
                      & \% Active neur. & 63 & 63 & 63 & 63 & 59 & 59 & 59 & 59 & 63 & 63 & 63 & 63  \\

\cmidrule(lr){1-6}\cmidrule(lr){7-10}\cmidrule(lr){11-14}

\multirow{2}{*}{PLANTS} & Accuracy & 0.408 & 0.403 & 0.343 & \textbf{0.479} & 0.392 & 0.420 & 0.313 &  \textbf{0.482} & 0.393 & 0.411 & 0.278 & \textbf{0.491}  \\
                      & \% Active neur. & 45 & 45 & 45 & 45 & 37 & 37 & 37 & 37 & 41 & 41 & 41 & 41 \\
\bottomrule
\end{tabular}}
}
\end{table*}

Second, Table \ref{tab:mean-cnn-evodeeptlpruning} shows the results for the \begin{idea2}evolution\end{idea2} of models with two layers, where \textbf{First Layer} indicates the cases of the \begin{idea2}evolution\end{idea2} of the first layer and \textbf{Second Layer} describes the cases of the second layer. Results point out that \proposal outperforms most of the methods in all the models and datasets. This case presents similar results as the one layer case because only in the PAINTING dataset \proposal has a lower performance in relation to the literature methods. As a result of that, \proposal's robustness in performance over the literature methods has been shown in one-layer and two-layer networks.

Lastly, we compare the execution times for all the models. Evolutionary approaches are known to converge slowly in highly-dimensional search spaces\begin{idea},\end{idea} as the one tackled in this paper. For that reason, in this section\begin{idea},\end{idea} we also want to compare the required time of \proposal and the other traditional approaches. Table \ref{tab:times-evodeeptlpruning-literature} shows the time in seconds for each \begin{idea}model\end{idea}. From these results, in terms of computational efficiency, our method suffers from the convergence slowness derived from the exploration of large search spaces. 

To summarize, in this section we have fairly compared \proposal to other well-known pruning methods, such as weight pruning and neuron pruning, guided by different pruning techniques. \proposal is distinguished from other pruning methods due to the fact that they are advocate for shrinking the through their pruning process, but with an admissible decrease of the accuracy. Although our model is slower in terms of execution time, it scores higher accuracy levels than those of traditional pruning counterparts. Therefore, we conclude that \proposal excels at determining which parameters to tune in neural networks with imported knowledge from other related tasks. 

\begin{table*}[htp!]
\centering
\caption{Times in seconds per run of \proposal against efficient CNN pruning methods with one and two layers models.}
\label{tab:times-evodeeptlpruning-literature}
\resizebox{\textwidth}{!}{%
{\begin{tabular}{crrrrrrrrrrr}
\toprule
& \multicolumn{6}{c}{\textbf{One Layer}} & \multicolumn{5}{c}{\textbf{Two Layers}}\\

\cmidrule(lr){3-6}\cmidrule(lr){7-12}

\multirow{2}{*}{Dataset} &  & \multirow{2}{*}{Weight} & Poly. & \multirow{2}{*}{Neuron} & \multirow{2}{*}{\proposal} & \multirow{2}{*}{Weight}& Poly.  & \multirow{2}{*}{Neuron} & \proposal & \proposal & \proposal \\
 & & & Decay &  &  &  & Decay &  & Layer 1 & Layer 2 & Both Layers\\
\cmidrule(lr){1-6}\cmidrule(lr){7-12}
SRSMAS & & 1,995 & 2125 & 1,995 & 34,510 & 2,395 & 2,545 & 2,398 & 34,856  & 38,731 & 55,596\\
RPS &  & 1,674 & 1,893 & 1,674 & 19,851 & 1,229 & 1,379 & 1,229 & 15,758 & 18,790 & 28,774\\
LEAVES & & 2,425 & 2,560 & 2,425 & 35,243 & 2,430 & 2,565 & 2,430 & 37,561 & 57,695 & 63,897\\
PAINTING & & 1,386 & 1,508 & 1,386 & 61,734 & 2,903 & 3,243 & 2,903 & 46,856 & 55,414 & 81,913 \\
CATARACT & & 594 & 627 & 584 & 6,768 &  449 & 473 & 449 & 7,392 & 8,529 & 12,350 \\
PLANTS &  & 298 & 407 & 298 & 28,456 &  270 & 370 & 270 & 21,788 & 22,235 & 35,998 \\
\bottomrule
\end{tabular}}
}
\end{table*}

\subsection{Answering RQ4: Feature Selection} \label{sec:rd-fs}

The RQ4 establishes the dichotomy of choosing pruning or feature selection for the given problem. For that reason, in this section we analyze the FS model by conducting the same group of experiments of the previous sections, to compare it against \proposal to decide which one scores best among them. The FS scheme is a particular case of \proposal if only one fully-connected layer composes the configuration of the network and the pruning and GA are focused on the extracted features of the ResNet-50 model.

Table \ref{tab:fs-ga} shows the results for this model against the reference methods. This case follows the same similarities of the previous ones, as FS obtains the best average results for all the datasets.

\begin{table}[H]
\centering
\caption{Average results for Fetaure Selection against non pruning methods.}
\label{tab:fs-ga}
%\resizebox{\columnwidth}{!}{%
%{
\begin{tabular}{ccccc}
  \toprule
  % Cambiar orden del nombre
    \multirow{2}{*}{Dataset} & \multirow{2}{*}{Measure} & \multirow{1}{*}{Not} &   \multirow{1}{*}{Best} & Feature \\
    &  & \multirow{1}{*}{Pruned} & \multirow{1}{*}{Fixed} & Selection \\
    \midrule
    \multirow{2}{*}{SRSMAS} & Accuracy & 0.832 &  0.866 &  \textbf{0.884} \\
                          & \% Active neur. & 100  & 20 & 60 \\
    \midrule
    \multirow{2}{*}{RPS} & Accuracy & 0.938 & 0.938 & \textbf{0.985} \\
                       & \% Active neur. & 100 & 40 & 45 \\
    \midrule
    \multirow{2}{*}{LEAVES} & Accuracy & 0.923 & 0.927 &  \textbf{0.943} \\
                          & \% Active neur. & 100 & 80 & 59  \\
    \midrule
    \multirow{2}{*}{PAINTING} & Accuracy& 0.939 & 0.945 &  \textbf{0.958}  \\
                              & \% Active neur. & 100 & 60 & 55  \\
    \midrule    
    \multirow{2}{*}{CATARACT} & Accuracy & 0.703 & 0.719 &  \textbf{0.747}  \\
                          & \% Active neur. & 100  & 70 & 55  \\
    \midrule    
    \multirow{2}{*}{PLANTS} & Accuracy & 0.432 & 0.432 &  \textbf{0.472} \\
                          & \% Active neur. & 100 & 10 & 68  \\
    \midrule
\end{tabular}%}}
\end{table}

Similarly to the previous sections, we have also compared this model with the CNN pruning methods with only one layer, as shown in Table \ref{tab:fs-ga-cnn}. In this case, in four out of six datasets the Feature Selection outperforms these methods, but in LEAVES and PAINTING  Weight and Polynomial Decay perform better than our model.

\begin{table}[!h]
\centering
\caption{Average results of Feature Selection against efficient CNN pruning methods.}
\label{tab:fs-ga-cnn}
%\resizebox{\columnwidth}{!}{%
%{
\begin{tabular}{llcccc}
\toprule
& & \multicolumn{4}{c}{\textbf{\begin{idea}
 Feature Selection
\end{idea}}}\\
\cmidrule(lr){3-6}

\multirow{2}{*}{Dataset} & \multirow{2}{*}{Measure} & \multirow{2}{*}{Weight} & Poly. &  \multirow{2}{*}{Neuron} & Feature \\
& & & Decay &  & Selection \\

\cmidrule(lr){1-6}%\cmidrule(lr){7-10}

\multirow{2}{*}{SRSMAS} & Accuracy & 0.841 & 0.878 & 0.802 & \textbf{0.884} \\
   & \% Active neur. & 60 & 60 & 60 & 60 \\

\cmidrule(lr){1-6}%\cmidrule(lr){7-10}

\multirow{2}{*}{RPS} & Accuracy & 0.913 & 0.926 & 0.869 & \textbf{0.985} \\
& \% Active neur. & 45 & 45 & 45 & 45 \\

\cmidrule(lr){1-6}%\cmidrule(lr){7-10}

\multirow{2}{*}{LEAVES} &  Accuracy & \textbf{0.947} & 0.940 & 0.946 & 0.943\\
   & \% Active neur. & 59 & 59 & 59 & 59 \\

\cmidrule(lr){1-6}%\cmidrule(lr){7-10}

\multirow{2}{*}{PAINTING} &  Accuracy & 0.962 & \textbf{0.968} & 0.883 & 0.958\\
   & \% Active neur. & 55 & 55 & 55 & 55 \\

\cmidrule(lr){1-6}%\cmidrule(lr){7-10}

\multirow{2}{*}{CATARACT} & Accuracy & 0.696 & 0.689 & 0.687 & \textbf{0.747} \\
   & \% Active neur. & 55 & 55 & 55 & 55 \\

\cmidrule(lr){1-6}%\cmidrule(lr){7-10}

\multirow{2}{*}{PLANTS} & Accuracy & 0.421 & 0.317 & 0.402 & \textbf{0.472} \\
   & \% Active neur. & 68 & 68 & 68 & 68\\
\bottomrule
\end{tabular}%}}
\end{table}

% conclusión sobre esta seccion
In this section\begin{idea},\end{idea} we have compared our FS scheme against reference models and efficient pruning methods published in the literature. The results shed light over the benefits of this model as it is also able to achieve a great performance over the reference models and also, in most cases, against the CNN pruning methods.

The global results for \proposal and its different versions are presented in Table \ref{tab:final-comparison}. The rows show the achieved accuracy and the percentage of improvement in relation to the best reference models for each model.

\begin{table}[!h]
\centering
\caption{Results and percentage of improvement for each version of \proposal in relation to each reference model.}
\label{tab:final-comparison}
%\resizebox{\textwidth}{!}{%
%{
\begin{tabular}{ccccc}
  \toprule
  % Cambiar orden del nombre
    \multirow{2}{*}{Dataset} & \multirow{2}{*}{Measure} & Pruning Model &  Pruning Model & Feature \\
    &   & One Layer & Both Layers & Selection \\
    \midrule
    \multirow{2}{*}{SRSMAS} & Accuracy & \textbf{0.885} & \textbf{0.885} & 0.884 \\
                          & \% Improvement & 1.9 & 2.2 & 1.8 \\
    \midrule
    \multirow{2}{*}{RPS} & Accuracy & 0.954 & 0.978 & \textbf{0.985} \\
                       & \% Improvement & 1.6 & 3.2 & 4.7 \\
    \midrule
    \multirow{2}{*}{LEAVES} & Accuracy & 0.935 & 0.936 &  \textbf{0.943} \\
                          & \% Improvement & 0.8 & 0.2 & 1.6 \\
    \midrule
    \multirow{2}{*}{PAINTING} & Accuracy& 0.951 & 0.953 &  \textbf{0.958}  \\
                              & \% Improvement & 0.6 & 0.4 & 1.3  \\
    \midrule    
    \multirow{2}{*}{CATARACT} & Accuracy & 0.732 & 0.746 &  \textbf{0.747}  \\
                          & \% Improvement & 1.3 & 1.1 & 2.8 \\
    \midrule    
    \multirow{2}{*}{PLANTS} & Accuracy & 0.480 & \textbf{0.491} &  0.472 \\
                          & \% Improvement & 4.8 & 2.5 & 4.0 \\
    \midrule
\end{tabular}%}}
\end{table}

Reviewing the results of \proposal and FS, we confirm that FS is the best model \begin{idea},\end{idea} as it obtains the best accuracy levels in four out of six datasets. Furthermore, the pruning of both layers carried out by \proposal also attains very notable performance levels.

Moreover, if we consider the \begin{idea2}evolution\end{idea2} using the pruning model, the \begin{idea2}evolution\end{idea2} of both layers yields the best results in terms of mean accuracy for each dataset. However, when comparing pruning and FS, the latter has more robust models: it achieves the best performance in four datasets\begin{idea},\end{idea} and it is also shown in the improvement percentage for each dataset.

In conclusion, it is shown with empirical evidence that pruning can be done by \begin{idea2}evolving\end{idea2} the fully-connected layers, specifically, by evolving their neurons to get the fittest configuration that reports an improvement of the network performance. An evolutionary feature selection based on the extracted features also achieves a great network performance, both in improving the accuracy and in reducing its complexity.

\begin{idea}

\subsection{Answering RQ5: Comparing different \proposal with different feature extractors} \label{sec:rd-extractors}

This section is devised to formally compare \proposal with different networks that serve as feature extractor for each dataset to analyze. CNNs have shown their capability to overcome different and diverse classification problems by learning visual features that best correlate with the target variable of the task at hand. Transferring this knowledge to other problems with a similar domain, which is what TL stands for, also helps to the capability of generalization of the model that is devised for the target task, specially when the volume of data for that task is low.

In previous sections, we have observed that the combination of \proposal with ResNet-50 has improved both reference models and pruning methods of the literature. However, in this section, we explore two other feature extractors and assess whether such performance gaps prevail. To determine performance gains of \proposal when using these alternative feature extractors, both are tested over the two best performing scenarios of \proposal, namely pruning both layers and feature selection. Moreover, we also include in the comparison these feature extractors with pruning algorithms from the literature, following the same experimental procedure described in preceding sections.

The chosen feature extractors are DenseNet-121 and VGG-19. The experiments with these networks have been carried out in the same conditions that the previous ones have been done. Table \ref{tab:comparison-networks-both} shows the comparison of these networks against the reference models based on fully-connected layers.

% analisis de los resultados
\begin{table*}[h!]
\centering
\caption{
\begin{idea}Average results of \proposal with different networks \begin{idea2}evolving\end{idea2} two consecutive layers against non pruning methods.\end{idea}
}
\label{tab:comparison-networks-both}
\resizebox{\textwidth}{!}{%
{\begin{tabular}{llccccccccc}
\toprule
& & \multicolumn{3}{c}{\textbf{ResNet-50}} & \multicolumn{3}{c}{\textbf{DenseNet-121}} & \multicolumn{3}{c}{\textbf{VGG19}} \\
\cmidrule(lr){3-5}\cmidrule(lr){6-8}\cmidrule(lr){9-11}

\multirow{2}{*}{Dataset} & \multirow{2}{*}{Measure} 
& \multirow{1}{*}{Not} &   \multirow{1}{*}{Best Fixed} & \multirow{1}{*}{EvoPrune} 
& \multirow{1}{*}{Not} &   \multirow{1}{*}{Best Fixed} & \multirow{1}{*}{EvoPrune} 
& \multirow{1}{*}{Not} &   \multirow{1}{*}{Best Fixed} & \multirow{1}{*}{EvoPrune} \\

&  & \multirow{1}{*}{Pruned} & \multirow{1}{*}{Both}& \multirow{1}{*}{DeepTL} 
& \multirow{1}{*}{Pruned} & \multirow{1}{*}{Both}& \multirow{1}{*}{DeepTL}
& \multirow{1}{*}{Pruned} & \multirow{1}{*}{Both}& \multirow{1}{*}{DeepTL} \\

\cmidrule(lr){1-5}\cmidrule(lr){6-8}\cmidrule(lr){9-11}

\multirow{2}{*}{SRSMAS} & Accuracy & 0.858 &  0.863 &  \textbf{0.885} & 0.861 & 0.881 & \textbf{0.890} & 0.837 & 0.853 & \textbf{0.885} \\
                        & \% Active neur. & 100 & 50 & 64 & 100 & 50 & 72 & 100 & 85 & 55 \\

\cmidrule(lr){1-5}\cmidrule(lr){6-8}\cmidrule(lr){9-11}

\multirow{2}{*}{RPS} & Accuracy & 0.922 &  0.946 & \textbf{0.978} & 0.704 & 0.723 & \textbf{0.754} & 0.814 & 0.879 & \textbf{0.922} \\
                     & \% Active neur. & 100 & 90 & 12 & 100 & 70 & 43 & 100 & 40 & 50 \\

\cmidrule(lr){1-5}\cmidrule(lr){6-8}\cmidrule(lr){9-11}

\multirow{2}{*}{LEAVES} & Accuracy & 0.919 & 0.934 &  \textbf{0.936} & 0.896 & 0.904 & \textbf{0.915} & 0.903 & 0.911 & \textbf{0.917} \\
                        & \% Active neur. & 100 & 15 & 34 & 100 & 60 & 39 & 100 & 50 & 68 \\

\cmidrule(lr){1-5}\cmidrule(lr){6-8}\cmidrule(lr){9-11}

\multirow{2}{*}{PAINTING} & Accuracy & 0.939 & 0.949 &  \textbf{0.953} & 0.940 & 0.943 & \textbf{0.947} & 0.923 & 0.938 & \textbf{0.945} \\
                          & \% Active neur. & 100  & 40 & 51 & 100 & 40 & 67 & 100 & 70 & 34 \\

\cmidrule(lr){1-5}\cmidrule(lr){6-8}\cmidrule(lr){9-11}

\multirow{2}{*}{CATARACT} & Accuracy & 0.703 & 0.735 &  \textbf{0.746} & 0.694 & 0.727 & \textbf{0.741} & 0.661 & 0.727 & \textbf{0.759} \\
                          & \% Active neur. & 100  & 85 & 63 & 100 & 55  & 51 & 100 & 45 & 42\\

\cmidrule(lr){1-5}\cmidrule(lr){6-8}\cmidrule(lr){9-11}
    
\multirow{2}{*}{PLANTS} & Accuracy & 0.402 & 0.466 &  \textbf{0.491} & 0.411 & 0.428 & \textbf{0.456} & 0.292 & 0.364 & \textbf{0.374} \\
                        & \% Active neur. & 100  & 55 & 41 & 100 & 55 & 78 & 100 & 50 & 64 \\

\bottomrule

\end{tabular}}
}
\end{table*}

The results from the previous table show the ability of \proposal to adapt itself to several feature extractors. For both DenseNet-121 and VGG-19, it improves the reference models. Taking a deep look at the three networks, ResNet-50 is the best of them, as it has the best improvement over several datasets. Nonetheless, the straight conclusion which is derived from these experiments is that \proposal is able to adapt to different feature extractors and datasets.

% tabla de comparación vs los modelos de la literatura
Once we have seen that \proposal has achieved better results than the reference models, now we inspect the performance of different pruning methods from the literature. For that reason, we compare the model which prunes two consecutive layers against its similar models from the pruning methods. Table \ref{tab:mean-cnn-evoprune-both} shows the comparison of the three networks against these methods. Note that, for this comparison, we have used the best to models of the CNN pruning methods: Weights and Polynomial Decay. This holds for the rest of this section.

\begin{table*}[h!]
\centering
\caption{\begin{idea}Average results of \proposal against efficient CNN pruning methods for pruning consecutive layers. \end{idea}}
\label{tab:mean-cnn-evoprune-both}
\resizebox{\textwidth}{!}{%
{\begin{tabular}{llcccccccccc}
\toprule
& & \multicolumn{3}{c}{\textbf{Both Layers - ResNet50}} & \multicolumn{3}{c}{\textbf{Both Layers - DenseNet-121}} &\multicolumn{3}{c}{\textbf{Both Layers - VGG19}}\\
\cmidrule(lr){3-5}\cmidrule(lr){6-8}\cmidrule(lr){9-11}

\multirow{2}{*}{Dataset} & \multirow{2}{*}{Measure} & \multirow{2}{*}{Weight} & Poly. & EvoPrune & \multirow{2}{*}{Weight} & Poly. & EvoPrune & \multirow{2}{*}{Weight} & Poly.  & EvoPrune \\
& & & Decay & DeepTL & & Decay & DeepTL & & Decay & DeepTL \\

\cmidrule(lr){1-5}\cmidrule(lr){6-8}\cmidrule(lr){9-11}

\multirow{2}{*}{SRSMAS} & Accuracy & 0.845 & 0.847 & \textbf{0.885} & 0.862 & 0.865 & \textbf{0.890} & \textbf{0.933} & 0.869 & 0.885  \\
   & \% Active neur. & 64 & 64 & 64 & 72 & 72 & 72 & 55 & 55 & 55 \\

\cmidrule(lr){1-5}\cmidrule(lr){6-8}\cmidrule(lr){9-11}

\multirow{2}{*}{RPS} & Accuracy & 0.694 & 0.899 & \textbf{0.978} & 0.815 & \textbf{0.817} & 0.754 & 0.803 & 0.842 & \textbf{0.922} \\
& \% Active neur. & 12 & 12 & 12 & 43 & 43 & 43 & 50 & 50 & 50 \\

\cmidrule(lr){1-5}\cmidrule(lr){6-8}\cmidrule(lr){9-11}

\multirow{2}{*}{LEAVES} & Accuracy & 0.911 & 0.925 & \textbf{0.936} & \textbf{1.000} & 0.907 & 0.915 & \textbf{1.000} & 0.916 & 0.917  \\
& \% Active neur. & 34 & 34 & 34 & 39 & 39 & 39 & 68 & 68 & 68  \\

\cmidrule(lr){1-5}\cmidrule(lr){6-8}\cmidrule(lr){9-11}

\multirow{2}{*}{PAINTING} & Accuracy & 0.934 & 0.940 & \textbf{0.953} & 0.897 & 0.901 & \textbf{0.947} & 0.923 & 0.922 & \textbf{0.945}   \\
    & \% Active neur. & 51 & 51 & 51 & 67 & 67 & 67 & 34 & 34 & 34 \\

\cmidrule(lr){1-5}\cmidrule(lr){6-8}\cmidrule(lr){9-11}

\multirow{2}{*}{CATARACT} & Accuracy & 0.686 & 0.696 & \textbf{0.746} & 0.587 & 0.593 & \textbf{0.741} & 0.661 & 0.686 & \textbf{0.759}  \\
                      & \% Active neur. & 63 & 63 & 63 & 51 & 51 & 51 & 42 & 42 & 42 \\

\cmidrule(lr){1-5}\cmidrule(lr){6-8}\cmidrule(lr){9-11}

\multirow{2}{*}{PLANTS} & Accuracy & 0.393 & 0.411 & \textbf{0.491} & 0.251 & 0.249 & \textbf{0.456} & 0.284 & 0.292 & \textbf{0.374} \\
                      & \% Active neur. & 41 & 41 & 41 & 78 & 78 & 78 & 64 & 64 & 64 \\
\bottomrule
\end{tabular}}
}
\end{table*}

The results shown in Table \ref{tab:mean-cnn-evoprune-both} give rise to interesting insights. To begin with, the first three columns are related to ResNet-50, which have a great performance over these reference models, and we know it from the previous sections. However, DenseNet and VGG are totally new in this kind of experiments. The reality is that both of these networks improve the CNN pruning methods when they are applied to fully-connected layers in most cases. Only in a few experiments are better than obtained by our proposal.

A global vision of these experiments suggests that DenseNet and VGG, just like ResNet, contribute to the discovery of pruned neural networks that maximize accuracy and reduce the number of active neurons. Moreover, these results verify that \proposal is able to achieve for different networks better results than reference and efficient CNN pruning methods when the pruning is made in two consecutive layers. Thus, we have shown the adaption ability of \proposal for this case using several networks (ResNet, DenseNet, and VGG). 

The following scenario is the feature selection model which derives from \proposal. This scenario, which encourages the pruning of the features extracted by the pre-trained network, has yielded the best results so far. Next, Table \ref{tab:comparison-networks-fs} shows the comparison of these networks when feature selection is performed.

\begin{table*}[h!]
\centering
\caption{\begin{idea}Average results for Feature Selection with different networks against non-pruning methods.\end{idea}}
\label{tab:comparison-networks-fs}
\resizebox{\textwidth}{!}{%
{\begin{tabular}{llccccccccc}
\toprule
& & \multicolumn{3}{c}{\textbf{ResNet-50}} & \multicolumn{3}{c}{\textbf{DenseNet-121}} & \multicolumn{3}{c}{\textbf{VGG19}} \\
\cmidrule(lr){3-5}\cmidrule(lr){6-8}\cmidrule(lr){9-11}

\multirow{2}{*}{Dataset} & \multirow{2}{*}{Measure} 
& \multirow{1}{*}{Not} &   \multirow{1}{*}{Best} & \multirow{1}{*}{Feature} 
& \multirow{1}{*}{Not} &   \multirow{1}{*}{Best} & \multirow{1}{*}{Feature} 
& \multirow{1}{*}{Not} &   \multirow{1}{*}{Best} & \multirow{1}{*}{Feature} \\

&  & \multirow{1}{*}{Pruned} & \multirow{1}{*}{Fixed}& \multirow{1}{*}{Selection} 
& \multirow{1}{*}{Pruned} & \multirow{1}{*}{Fixed}& \multirow{1}{*}{Selection}
& \multirow{1}{*}{Pruned} & \multirow{1}{*}{Fixed}& \multirow{1}{*}{Selection} \\

\cmidrule(lr){1-5}\cmidrule(lr){6-8}\cmidrule(lr){9-11}

\multirow{2}{*}{SRSMAS} & Accuracy & 0.832 &  0.866 &  \textbf{0.884} & 0.858 & 0.881 & \textbf{0.896} & 0.753 & 0.766 & \textbf{0.869} \\
                        & \% Active neur. & 100  & 20 & 60 & 100 & 60 & 68 & 100 & 40 & 87 \\

\cmidrule(lr){1-5}\cmidrule(lr){6-8}\cmidrule(lr){9-11}

\multirow{2}{*}{RPS} & Accuracy & 0.938 &  0.938 & \textbf{0.985} & 0.720 & 0.720 & \textbf{0.839} & 0.887 & 0.890 & \textbf{0.982} \\
                     & \% Active neur. & 100 & 40 & 45 & 100 & 100 & 48 & 100 & 60 & 53 \\

\cmidrule(lr){1-5}\cmidrule(lr){6-8}\cmidrule(lr){9-11}

\multirow{2}{*}{LEAVES} & Accuracy & 0.923 & 0.927 &  \textbf{0.943} & 0.896 & 0.902 & \textbf{0.921} & 0.852 & 0.876 & \textbf{0.924} \\
                        & \% Active neur. & 100  & 80 & 59 & 100 & 10 & 60 & 100 & 10 & 68 \\

\cmidrule(lr){1-5}\cmidrule(lr){6-8}\cmidrule(lr){9-11}

\multirow{2}{*}{PAINTING} & Accuracy & 0.939 & 0.945 &  \textbf{0.958} & 0.934 & 0.941 & \textbf{0.956} & 0.924 & 0.924 & \textbf{0.943} \\
                          & \% Active neur. & 100  & 60 & 55 & 100 & 90 & 63 & 100 & 100 & 77 \\

\cmidrule(lr){1-5}\cmidrule(lr){6-8}\cmidrule(lr){9-11}

\multirow{2}{*}{CATARACT} & Accuracy & 0.703 & 0.719 &  \textbf{0.747} & 0.669 & 0.702 & \textbf{0.787} & 0.628 & 0.661 & \textbf{0.765} \\
                          & \% Active neur. & 100  & 70 & 55 & 100 & 30 & 57 & 100 & 10 & 66 \\

\cmidrule(lr){1-5}\cmidrule(lr){6-8}\cmidrule(lr){9-11}
    
\multirow{2}{*}{PLANTS} & Accuracy & 0.432 & 0.432 &  \textbf{0.472} & 0.394 & 0.415 & \textbf{0.464} & 0.335 & 0.352 & \textbf{0.376} \\
                        & \% Active neur. & 100 & 10 & 68 & 100 & 90 & 67 & 100 & 40 & 66 \\

\bottomrule

\end{tabular}}
}
\end{table*}

% analisis de los resultados
In the previous sections, we have shown that \proposal is able to prune the extracted features derived from the network, and this model has reached the best results of \proposal. Table \ref{tab:comparison-networks-fs} shows that the pruning of the features that they have been extracted using different networks (DenseNet and VGG) also increases the performance of the networks, which is shown in the accuracy of these networks over the datasets. There are some cases in which non pruning methods have less active neurons, but their accuracy is lower than the models of \proposal. For that reason, the feature selection keeps being the best of \proposal models, as it has the best results so far.

% tabla de comparacion vs los modelos de la literatura
The next, and final, step is to check the performance of this feature selection model against efficient CNN pruning methods from the literature. In this section, we have checked that, for models which prune two consecutive layers, \proposal performs better than the pruning methods. Consequently, now we focus on this comparison, but in terms of the models which prune the extracted features. Table \ref{tab:mean-cnn-evoprune-fs} shows the results of this comparison for the different networks.

\begin{table*}[ht!]
\centering
\caption{\begin{idea}Average results of \proposal with different networks against efficient CNN pruning methods for feature selection models.\end{idea}}
\label{tab:mean-cnn-evoprune-fs}
\resizebox{\textwidth}{!}{%
{\begin{tabular}{llcccccccccc}
\toprule
& & \multicolumn{3}{c}{\textbf{Feature Selection - ResNet50}} & \multicolumn{3}{c}{\textbf{Feature Selection - DenseNet-121}} &\multicolumn{3}{c}{\textbf{Feature Selection - VGG19}}\\
\cmidrule(lr){3-5}\cmidrule(lr){6-8}\cmidrule(lr){9-11}

\multirow{2}{*}{Dataset} & \multirow{2}{*}{Measure} & \multirow{2}{*}{Weight} & Poly. & Feature & \multirow{2}{*}{Weight} & Poly. & Feature & \multirow{2}{*}{Weight} & Poly.  & Feature \\
& & & Decay & Selection & & Decay & Selection & & Decay & Selection \\

\cmidrule(lr){1-5}\cmidrule(lr){6-8}\cmidrule(lr){9-11}

\multirow{2}{*}{SRSMAS} & Accuracy & 0.841 & 0.878 & \textbf{0.884} & 0.869 & 0.868 & \textbf{0.896} & 0.826 & 0.825 & \textbf{0.869}  \\
   & \% Active neur. & 60 & 60 & 60 & 68 & 68 & 68 & 87 & 87 & 87 \\

\cmidrule(lr){1-5}\cmidrule(lr){6-8}\cmidrule(lr){9-11}

\multirow{2}{*}{RPS} & Accuracy & 0.913 & 0.926 & \textbf{0.985} & 0.675 & 0.699 & \textbf{0.839} & 0.981 & 0.834 & \textbf{0.982}\\
& \% Active neur. & 45 & 45 & 45 & 48 & 48 & 48 & 53 & 53 & 53 \\

\cmidrule(lr){1-5}\cmidrule(lr){6-8}\cmidrule(lr){9-11}

\multirow{2}{*}{LEAVES} & Accuracy & \textbf{0.947} & 0.940 & 0.943 & 0.858 & 0.891 & \textbf{0.921} & 0.904 & 0.843 & \textbf{0.924} \\
& \% Active neur. & 59 & 59 & 59 & 60 & 60 & 60 & 68 & 68 & 68  \\

\cmidrule(lr){1-5}\cmidrule(lr){6-8}\cmidrule(lr){9-11}

\multirow{2}{*}{PAINTING} & Accuracy & 0.962 & \textbf{0.968} & 0.958 & 0.937 & 0.934 & \textbf{0.956} & 0.928 & 0.928 & \textbf{0.943} \\
    & \% Active neur. & 55 & 55 & 55 & 63 & 63 & 63 & 77 & 77 & 77 \\

\cmidrule(lr){1-5}\cmidrule(lr){6-8}\cmidrule(lr){9-11}

\multirow{2}{*}{CATARACT} & Accuracy & 0.696 & 0.689 & \textbf{0.747} & 0.676 & 0.682 & \textbf{0.787} & 0.666 & 0.688 & \textbf{0.765} \\
                      & \% Active neur. & 55 & 55 & 55 & 57 & 57 & 57 & 66 & 66 & 66 \\

\cmidrule(lr){1-5}\cmidrule(lr){6-8}\cmidrule(lr){9-11}

\multirow{2}{*}{PLANTS} & Accuracy & 0.421 & 0.317 & \textbf{0.472} & 0.387 & 0.394 & \textbf{0.464} & 0.322 & 0.311 & \textbf{0.376}  \\
                      & \% Active neur. & 68 & 68 & 68 & 67 & 67 & 67 & 66 & 66 & 66 \\
\bottomrule
\end{tabular}}
}
\end{table*}

The results show, not only that ResNet-50 has a great performance (same results as previous sections), that both DenseNet and VGG outperform the pruning methods when applied to prune the features extracted from the networks. Both new networks show a better performance in all the datasets than the reference methods. For that reason, we conclude that the usage of \proposal with these three networks has proven the capability to perform better than the pruning methods.

% conclusion
In this section, we have compared ResNet-50 with other two networks in two different scenarios: pruning consecutive layers and pruning the extracted features from the networks. The experiments show that \proposal has proven its ability to adapt to other networks in both cases and has improved both reference models and pruning methods of the literature. For that reason, and in light of the results from the previous sections, we conclude that \proposal has shown the ability of facing diverse tasks, as \proposal has achieved a great performance when different networks are used either for pruning consecutive layer or pruning the features extracted from the network.
\end{idea}

\begin{idea}
\subsection{Answering RQ6: Analyzing the ability of \proposal to adapt to relevant classes and robustness.}
\label{sec:rd-generalization}

The purpose of this section is twofold. First, we want to analyze the goodness of \proposal when modeling varying problems. In this case, we want to see how it adapts to the different classes that make up the datasets so that it captures the relevance of each of them. For a given dataset, we analyze each of its classes to determine if \proposal is also able to have a good performance over it, and then, compare these results with the whole dataset.

The second objective of this section is the analysis of these results. Once \proposal has modelled each of the classes for a dataset at hand, we check the quality of the obtained results. For that reason, we must check that results are not affected by the stochasticity induced by the usage of a genetic algorithm at the core of the proposed \proposal. Recall that stochasticity implies that the output of the algorithm may not be the same, even with the same input. For that reason, our second objective is related to this factor, and we want to show that the effect of randomness has a low impact on \proposal, i.e., the good results do not depend on the randomness.

In order to measure the effect of randomness in the pruned networks evolved by \proposal, we resort to a similarity measure called Centered Kernel Alignment, CKA \citep{CKA}. CKA measures the similarity of trained neural networks, in compliance with several invariance properties that must be met for these particular computing structures (namely invariance to invertible linear transformation, invariance to orthogonal transformation and invariance to isotropic scaling). We compare the trained networks as a result of the application of \proposal. This comparison answers the question about the robustness of \proposal.

CKA is based on the Hilbert-Schmidt Independence Criterion, HSIC \citep{hsic}. It compares two matrices (\textbf{K} and \textbf{L}) and determines the level of independence between them, as it is shown in \ref{eq:cka}. In our case, these matrices are the structures that contain the weights of the trained neural networks. CKA takes a maximum value of 1 when the two inputs of CKA are the same matrix. The range of this measure is $[0,1]$. This means that both matrix are very similar (in that case because they are identical). Thus, if the CKA value is high, then both matrix are similar.

\begin{equation} \label{eq:cka}
    CKA(K,L) = \frac{HSIC(K,L)}{\sqrt{HSIC(K,K)HSIC(L,L)}}
\end{equation}\normalsize 

CKA is a measure which allows us make a double comparison. Note that this measure helps us to compare the genotype (chromosomes of the GA) against the phenotype (pruned networks of \proposal). The interpretation of this measure follows that if the similarity of the chromosomes is high, then the CKA value should be also high (close to one, which is its maximum). Five independent executions have been made of \proposal, so we have taken the output of each of them, and we have performed a double comparison based on this trained neural network, which is explained next:

\begin{enumerate}
    \item Comparison against the closest element (\textit{CKA$_{Closest}$}): we have the \textit{Output} element as the best one for an execution. Then, we calculate the Hamming distance of the best element with respect to all the elements evaluated in the evolutionary process of \proposal. The element with the smallest Hamming distance to the best one is denoted as \textit{Closest}. Then, \textit{Output} is compared against \textit{Closest}. The robustness of \proposal is tested in this comparison because high values of CKA when similar chromosomes are compared is essential, as it will show that the results are not due to randomness, but to the process that \proposal performs.
    
    \item Comparison against fully-connected, reference models (CKA$_{Ref}$): in this case, we compare \textit{Output} against a fully-connected network with all the neurons activated. This comparison sheds light on the ability of \proposal to learn which neurons are the best to solve the problem at hand. It also permits to explain the difference in terms of accuracy between the models that \proposal develops against the reference models. This value of CKA quantifies the differences in accuracy between our models and the reference models, as \proposal searches for the best neurons to remove the unnecessary ones, while reference models simply train the models without taking into account the neurons which should be removed.
\end{enumerate}

In this section, we select the best two models of \proposal for these experiments: pruning consecutive layers (both case) and pruning the features extracted from the networks (feature selection). Moreover, three datasets are considered in the experiments designed for this section: CATARACT, PAINTING and RPS. Lastly, we show different tables with the following structure: \textit{DATASET-Class}. This means that the mentioned \textit{DATASET} is analyzed without the class called \textit{Class}.

For the CKA comparison, we will show two groups of tables, one per each type of model (pruning consecutive layer and feature selection). Moreover, each table is composed of the problem at hand and the mean values of the Hamming distance and CKA averaged over five executions of \proposal. The rows of each table correspond to each \textit{DATASET-Class} and the four columns represent in pairs the previously explained comparisons, CKA$_{Closest}$ and CKA$_{Ref}$, as we show the mean Hamming distance and its corresponding CKA mean value for each of the comparisons.

\subsubsection{Analyzing the relevance of each class for a given dataset}

Given a dataset of $n$ classes, this approach performs a procedure that removes a whole class of the dataset and then, \proposal is applied with that remaining data both in pruning consecutive layers and in feature selection. As a result of that process, $n$ experiments are done for each dataset.

The structure of the these tables correspond with the usual structure of the rest of the paper, but now we show four columns. These columns represent, in pairs based on the model, the results of \proposal versus the reference model without pruning. The difference between the pairs of columns is the model at hand: pruning consecutive layers or feature selection. We note that for the first two columns, the \begin{idea2}evolution\end{idea2} process which lies in \proposal is made with 300 evaluations and two-layers networks, meanwhile for the last columns, the process of pruning the extracted features is made with only one-layer networks and 200 evaluations, i.e., under the same conditions as the experiments in the previous sections.

The first results show the CATARACT dataset under these conditions. Table \ref{tab:cataract-class} shows these results. We conclude from these results that Cataract class is the easiest class in the dataset, as both models struggle with that class (third row of the table), but they improve their results with it. However, the results of \proposal are better than the reference model. The number of active neurons at the end of the evolutionary process are reduced, in most cases, nearly by half. Moreover, the feature selection model is also able to decrease this number from its previous results (55\% of remaining active neurons) in some cases. The same conclusion can be drawn in relation to pruning consecutive layers, as the full dataset has a mean percentage of active neurons of 63\% and in three of fours cases this number is reduced. The results show that the pruning of the extracted features of the network, i.e., the feature selection which derives from \proposal, is the best approximation for this dataset.

% añadir las tablas como las del informe para CATARACT, RPS y PAINTING con both y feature selection
\begin{table}[H]
\centering
\caption{\begin{idea}Average results of CATARACT-Class with pruning consecutive layers and feature selection.\end{idea}}
\label{tab:cataract-class}
%\resizebox{\columnwidth}{!}{%
%{
\begin{tabular}{*{1}{c}*{5}{c}}
  \toprule
  % Cambiar orden del nombre
    \multirow{2}{*}{Dataset} & \multirow{2}{*}{Measure} & \multirow{1}{*}{\proposal} & No Pruning  & \multirow{1}{*}{\proposal} & No Pruning \\
    & & \multirow{1}{*}{Both} & Both & \multirow{1}{*}{Feature Selection} & Feature Selection \\
    \midrule
    CATARACT -  & Accuracy & \textbf{0.844} & 0.801 & \textbf{0.871} & 0.697 \\
    Retina  & \% Active neur. & 43 & 100 & 54 & 100 \\
    \midrule
    CATARACT - & Accuracy & \textbf{0.846} & 0.789 & \textbf{0.857} & 0.554 \\
    Glaucoma   & \% Active neur. & 44 & 100 & 63 & 100 \\
    \midrule
    CATARACT -  & Accuracy & \textbf{0.735} & 0.732 & \textbf{0.761} & 0.614  \\
    Cataract & \% Active neur. & 76 & 100 & 49 & 100 \\
    \midrule
    CATARACT - & Accuracy &  \textbf{0.833} & 0.805 & \textbf{0.843} & 0.655 \\
    Normal   & \% Active neur. & 53 & 100 & 57 & 100 \\
    \midrule
    CATARACT - & Accuracy &  \textbf{0.746} & 0.703 & \textbf{0.747} & 0.703 \\
    Full   & \% Active neur. & 63 & 100 & 55 & 100 \\
    \midrule
\end{tabular}
%}}
\end{table}

We focus now on the following dataset, RPS. Table \ref{tab:rps-class} shows the results of these experiments. Results show that the \textit{Paper} class makes an easier dataset, as all the approaches reach the maximum accuracy. The other two experiments show that \proposal achieves a better performance to the reference models. For pruning consecutive layers, the number of remaining active neurons is higher in comparison with the full dataset, but in the feature selection model this number is very similar or even for RPS-Scissors is lower.

\begin{table}[htp]
\centering
\caption{\begin{idea}Average results of general RPS pruning consecutive layers and feature selection.\end{idea}}
\label{tab:rps-class}
%\resizebox{\columnwidth}{!}{%
%{
\begin{tabular}{*{1}{c}*{5}{c}}
  \toprule
  % Cambiar orden del nombre
\multirow{2}{*}{Dataset} & \multirow{2}{*}{Measure} & \multirow{1}{*}{\proposal} & No Pruning  & \multirow{1}{*}{\proposal} & No Pruning \\
    & & \multirow{1}{*}{Both} & Both & \multirow{1}{*}{Feature Selection} & Feature Selection \\
    \midrule
    RPS -  & Accuracy & \textbf{1.000} & 1.000 & \textbf{1.000} & 1.000 \\
    Paper  & \% Active neur. & 51 & 100 & 51 & 100 \\
    \midrule
    RPS - & Accuracy & \textbf{0.985} & 0.979 & \textbf{1.000} & 0.996 \\
    Rock   & \% Active neur. & 51 & 100 & 49 & 100 \\
    \midrule
    RPS -  & Accuracy & \textbf{0.994} & 0.955 & \textbf{1.000} & 0.955 \\
    Scissors & \% Active neur. & 33 & 100 & 23 & 100 \\
    \midrule
    RPS & Accuracy & 0.978 & 0.922 & 0.985 & 0.938 \\
    Full     & \% Active neur. & 12 & 100 & 45 & 100 \\  
    \midrule
\end{tabular}
%}}
\end{table}

The last considered dataset is PAINTING. Table \ref{tab:painting-class} shows the results for each of the experiments which have been made for this dataset. The table shows that both of \proposal models are constantly achieving better results than the reference models. However, the difference between the accuracy is higher in the feature selection models than the pruning consecutive layers. Taking into consideration the remaining active neurons, the feature selection has a similar degree of pruning in relation with the experiments which have been carried out for the full dataset experiments (55\%). Similar conclusion can be drawn for the other model, as the mean percentage of active neuron for pruning consecutive layers is 51\% and we have models with fewer active neurons, but also with higher percentage.

\begin{table}[H]
\centering
\caption{\begin{idea}Average results of general PAINTING pruning consecutive layers and feature selection.\end{idea}}
\label{tab:painting-class}
%\resizebox{\columnwidth}{!}{%
%{
\begin{tabular}{*{1}{c}*{5}{c}}
  \toprule
  % Cambiar orden del nombre
\multirow{2}{*}{Dataset} & \multirow{2}{*}{Measure} & \multirow{1}{*}{\proposal} & No Pruning  & \multirow{1}{*}{\proposal} & No Pruning \\
    & & \multirow{1}{*}{Both} & Both & \multirow{1}{*}{Feature Selection} & Feature Selection \\
    \midrule
    PAINTING - & Accuracy & \textbf{0.942} & 0.932 & \textbf{0.947} & 0.857 \\
    Sculpture  & \% Active neur. & 64 & 100 & 53 & 100 \\
    \midrule
    PAINTING - & Accuracy & \textbf{0.959} & 0.946 & \textbf{0.961} & 0.820 \\
    Painting   & \% Active neur. & 42 & 100 & 50 & 100 \\
    \midrule
    PAINTING -  & Accuracy & \textbf{0.942} & 0.920 & \textbf{0.949} & 0.835 \\
    Iconography & \% Active neur. & 30 & 100 & 57 & 100 \\
    \midrule
    PAINTING - & Accuracy & \textbf{0.979} & 0.974 & \textbf{0.979} & 0.883 \\
    Engraving  & \% Active neur. & 74 & 100 & 56 & 100 \\
    \midrule    
    PAINTING - & Accuracy & \textbf{0.994} & 0.989 & \textbf{0.996} & 0.944 \\
    Drawings   & \% Active neur. & 57 & 100 & 57 & 100 \\
    \midrule
    PAINTING & Accuracy & 0.953 & 0.939 & 0.958 & 0.939 \\
    Full     & \% Active neur. & 51 & 100 & 55 & 100 \\
    \midrule
\end{tabular}
%}}
\end{table}

% conclusion de estos experimentos.
These experiments shed light on a conclusion that it is not far from the previous sections. The feature selection model, which performs the pruning of the extracted features of the network, constitutes the best model for all the experiments, as it has the most difference between \proposal and reference models. Nonetheless, the effect of pruning consecutive layers is also positive, as it is shown in the results of the previous tables.

The next part of this section is crucial to determine the robustness of \proposal. Moreover, the differences in accuracy of the previous tables are going to be explained in the following tables. The key element of the comparison is the CKA measure and the Hamming distance of the solutions. The combination of these values determine the key points that we have discussed at the beginning of this section.

Therefore, we are showing the CKA tables for each experiment to perform the commented double comparison in this section. These tables have a different structure from the last tables, so we explain how to interpret them. Both of them present a similar structure, but the table which shows the results for pruning consecutive layers has another column. This column shows the value of the CKA measure for the layer at hand. In the case of the prune of the extracted feature of the network (feature selection), as they only has one layer, this column is not required.

We show this pair of tables for each \textit{DATASET-Class}. The first table shows the results of the feature selection model, and the second one presents the results of the pruning of consecutive layers. 

The composition of the tables for feature selection models is the following one. After the first column which show the dataset at hand, the next two columns represent the comparison among the phenotype and genotype of \proposal, i.e, chromosomes and pruned networks. Both mean Hamming distance of the five executions and CKA$_{Closest}$ are shown. The following group of two columns shows the other explained comparison of the reference models, which have all the neurons active. The metrics are the same as the previous case, but now the CKA (CKA$_{Ref}$) corresponds to the mean value from the best model to this reference model.

The composition of the table for the pruning of consecutive layers is similar to the previous case. However, another column is required for a more detailed explanation. This column gives information about the CKA value for the layer at hand. Due to the fact that we are comparing the whole chromosome, both Hamming distance values are common to both layers, but the CKA value is layer dependent. Then, in this table, we want to highlight that the best chromosome \proposal obtains good values of the CKA measure for each of the layers of the model. 

First, we show the CKA values for the CATARACT dataset in its four different cases of \textit{DATASET-Class}. Table \ref{tab:cataract-cka-fs} and Table \ref{tab:cataract-cka-both} show the results for feature selection and pruning consecutive layers, respectively. The results show for both models the robustness of \proposal because the mean CKA of the best versus its closest chromosome in the history is a value extremely close to 1, which is the maximum value (this value is reached when the best solution is compared to itself). 

Taking a more deep look at the results of the feature selection, we see that the Hamming distance is very low, which is a good result and also proves the robustness of \proposal. The second group of columns, in which the comparison is made against a model with all neurons active, we see that the Hamming distance is higher and this has an impact on the CKA value, which is higher. The conclusion which derives from these experiments is that \proposal learns to distinguish the valuable neurons which have an impact on the model. This CKA value is the explanation of the difference in accuracy in the previous experiments of this dataset between \proposal models and reference models.

% añadir las tablas de CKA para cada uno de los experimentos
\begin{table}[H]
\centering
\caption{\begin{idea}Comparison of the CKA measure for feature selection in CATARACT-Class.\end{idea}}
\label{tab:cataract-cka-fs}
%\resizebox{\columnwidth}{!}{%
%{
\begin{tabular}{*{1}{c}*{4}{c}}
  \toprule
  % Cambiar orden del nombre
    &  \multicolumn{4}{c}{\textbf{Feature Selection}}\\
    \cmidrule(lr){2-5}
    \multirow{2}{*}{Dataset} & \multirow{1}{*}{Hamming Distance} & \multirow{2}{*}{CKA$_{Closest}$}  & \multirow{1}{*}{Hamming Distance} & \multirow{2}{*}{CKA$_{Ref}$} \\
    & \proposal & & No Pruned Model &  \\
    \midrule
    CATARACT - & \multirow{2}{*}{0.005} & \multirow{2}{*}{0.981} & \multirow{2}{*}{0.457} & \multirow{2}{*}{0.283} \\
    Retina  &  &  &  &  \\
    \midrule
    CATARACT -  & \multirow{2}{*}{0.002} & \multirow{2}{*}{0.991} & \multirow{2}{*}{0.372} & \multirow{2}{*}{0.376} \\
    Glaucoma    &  &  &  &  \\
    \midrule
    CATARACT -   & \multirow{2}{*}{0.001} & \multirow{2}{*}{0.994} & \multirow{2}{*}{0.514} & \multirow{2}{*}{0.200} \\
    Cataract  &  &  &  &  \\
    \midrule
    CATARACT - &  \multirow{2}{*}{0.011} & \multirow{2}{*}{0.961} & \multirow{2}{*}{0.423} & \multirow{2}{*}{0.250} \\
    Normal    &  &  &  &  \\
    \midrule
\end{tabular}
%}}
\end{table}

The results that they are shown in Table \ref{tab:cataract-cka-both} confirm the robustness of \proposal. This insight is the same as in the previous table: low values of Hamming distance in the best chromosomes of \proposal implies high values of CKA in the closest element. Moreover, the Hamming distance from the elements of pruning consecutive layers is higher than in the other case, but the CKA values are also higher. This is a fair result because the difference in accuracy between \proposal and the reference models is lower than in the other case. Note that the class Cataract from this dataset is the class with the fewest gap in accuracy, and this is shown in its CKA value. The Glaucoma class is the opposite of Cataract, and the CKA value is lower. In all the cases, \proposal confirms the ability to learn the neurons that they are indispensable to achieve a greater performance, and that is the main difference between the reference models.

\begin{table}[H]
\centering
\caption{\begin{idea}Comparison of the CKA measure for pruning consecutive layers in CATARACT-Class.\end{idea}}
\label{tab:cataract-cka-both}
\resizebox{\columnwidth}{!}{%
{
\begin{tabular}{*{1}{c}*{5}{c}}
  \toprule
  % Cambiar orden del nombre
    &  & \multicolumn{3}{c}{\textbf{Pruning consecutive layers}}\\
    \cmidrule(lr){2-6}
    \multirow{2}{*}{Dataset} & \multirow{2}{*}{\# Layer} & \multirow{1}{*}{Hamming Distance} & \multirow{2}{*}{CKA$_{Closest}$}  & \multirow{1}{*}{Hamming Distance} & \multirow{2}{*}{CKA$_{Ref}$} \\
    & & \proposal & & No Pruned Model &  \\
    \midrule
    CATARACT -  & Layer 1 & \multirow{2}{*}{0.020} & 0.980 & \multirow{2}{*}{0.567} & 0.685 \\
    Retina      & Layer 2 &  & 0.983 &  &  0.769 \\
    \midrule
    CATARACT -  & Layer 1 & \multirow{2}{*}{0.001} & 0.996 & \multirow{2}{*}{0.556} & 0.680 \\
    Glaucoma    & Layer 2 &  & 0.997 &  & 0.745 \\
    \midrule
    CATARACT -  & Layer 1 & \multirow{2}{*}{0.001} & 0.998 & \multirow{2}{*}{0.240} & 0.891 \\
    Cataract    & Layer 2 &  & 0.998 &  & 0.916 \\
    \midrule
    CATARACT -  & Layer 1 & \multirow{2}{*}{0.027} & 0.975 & \multirow{2}{*}{0.469} & 0.784 \\
    Normal      & Layer 2 &  & 0.979 & & 0.843\\
    \midrule
\end{tabular}
}}
\end{table}

The second dataset under analysis is RPS. Table \ref{tab:rps-cka-fs} shows the results of the feature selection models for RPS-Class. We see that the pair Hamming distance and CKA of the closest have a great result in two of the three cases. Moreover, the results of the other metrics achieve a great results, similarly to CATARACT-Class with this model. For that reason, we confirm that, for this model, \proposal is also able to learn the neurons that maximize the accuracy for the problem. 

A special case is RPS-Paper. When RPS does not have this class, the problem seems to be a very easy task, because all the models in the previous experiments for this dataset achieve the maximum accuracy. This is the only case in which the CKA for the best and its closest element is lower in comparison with the others. This is due to the fact that the problem at hand can be solved with many chromosomes, as they all have the maximum accuracy value, so the chromosomes might not be very similar, because the range of possible solutions is wide.

\begin{table}[H]
\centering
\caption{\begin{idea}Comparison of the CKA measure for feature selection in RPS-Class.\end{idea}}
\label{tab:rps-cka-fs}
%\resizebox{\columnwidth}{!}{%
%{
\begin{tabular}{*{1}{c}*{4}{c}}
  \toprule
  % Cambiar orden del nombre
    &  \multicolumn{4}{c}{\textbf{Feature Selection}}\\
    \cmidrule(lr){2-5}
    \multirow{2}{*}{Dataset} & \multirow{1}{*}{Hamming Distance} & \multirow{2}{*}{CKA$_{Closest}$}  & \multirow{1}{*}{Hamming Distance} & \multirow{2}{*}{CKA$_{Ref}$} \\
    & \proposal & & No Pruned Model &  \\
    \midrule
    RPS - & \multirow{2}{*}{0.199} & \multirow{2}{*}{0.416} & \multirow{2}{*}{0.485} & \multirow{2}{*}{0.224} \\
    Paper  &  &  &  &  \\
    \midrule
    RPS -  & \multirow{2}{*}{0.020} & \multirow{2}{*}{0.926} & \multirow{2}{*}{0.507} & \multirow{2}{*}{0.195} \\
    Rock    &  &  &  &  \\
    \midrule
    RPS -   & \multirow{2}{*}{0.040} & \multirow{2}{*}{0.863} & \multirow{2}{*}{0.766} & \multirow{2}{*}{0.104} \\
    Scissors  &  &  &  &  \\
    \midrule
\end{tabular}
%}}
\end{table}

Table \ref{tab:rps-cka-both} shows the pruning of consecutive layers that it is performed by \proposal. Similar conclusions are obtained from these results. First, we see that the Paper class has the same problem which appears in the previous table. However, the other two groups of experiments are harder to solve and this have been drawn in the CKA of the closest element, because both layers have a great value of this measure. In the counterpart, the CKA values for the reference also has a similar understanding, which belongs to the fact that \proposal pruning of consecutive layers learns the best neurons for both layers.

In overall, \proposal is also a robust model for this dataset both in feature selection and pruning consecutive layers, and it also proves that the difference in accuracy between our models and the reference models is stated in the CKA$_{Ref}$.

\begin{table}[H]
\centering
\caption{\begin{idea}Comparison of the CKA measure for pruning consecutive layers in RPS-Class.\end{idea}}
\label{tab:rps-cka-both}
\resizebox{\columnwidth}{!}{%
{
\begin{tabular}{*{1}{c}*{5}{c}}
  \toprule
  % Cambiar orden del nombre
      &  & \multicolumn{3}{c}{\textbf{Pruning consecutive layers}}\\
    \cmidrule(lr){2-6}
    \multirow{2}{*}{Dataset} & \multirow{2}{*}{\# Layer} & \multirow{1}{*}{Hamming Distance} & \multirow{2}{*}{CKA$_{Closest}$}  & \multirow{1}{*}{Hamming Distance} & \multirow{2}{*}{CKA$_{Ref}$} \\
    & & \proposal & & No Pruned Model &  \\
    \midrule
    RPS -    & Layer 1 & \multirow{2}{*}{0.217} & 0.725 & \multirow{2}{*}{0.486} & 0.764 \\
    Paper    & Layer 2 &  & 0.761 &  &  0.825\\
    \midrule
    RPS -    & Layer 1 & \multirow{2}{*}{0.032} & 0.977 & \multirow{2}{*}{0.492} & 0.757 \\
    Rock     & Layer 2 &  & 0.983 &  & 0.852 \\
    \midrule
    RPS -    & Layer 1 & \multirow{2}{*}{0.010} & 0.982 & \multirow{2}{*}{0.667} & 0.597 \\
    Scissors & Layer 2 &  & 0.974 & & 0.668 \\
    \midrule
\end{tabular}
}}
\end{table}

The last dataset in this section is PAINTING. The first table relates to the feature selection model of \proposal. Table \ref{tab:painting-cka-fs}, once more, shows that the lowest Hamming distance of the best chromosome when it is compared with its closest, brings high values of CKA. The conclusion is clear, \proposal is robust. Moreover, the values of CKA$_{Ref}$ are also a good estimation of how \proposal looks for the best neurons. Both CATARACT and PAINTING have lots of similarities in the feature selection model.

\begin{table}[H]
\centering
\caption{\begin{idea}Comparison of the CKA measure for feature selection in PAINTING-Class.\end{idea}}
\label{tab:painting-cka-fs}
%\resizebox{\columnwidth}{!}{%
%{
\begin{tabular}{*{1}{c}*{4}{c}}
  \toprule
  % Cambiar orden del nombre
    &  \multicolumn{4}{c}{\textbf{Feature Selection}}\\
    \cmidrule(lr){2-5}
    \multirow{2}{*}{Dataset} & \multirow{1}{*}{Hamming Distance} & \multirow{2}{*}{CKA$_{Closest}$}  & \multirow{1}{*}{Hamming Distance} & \multirow{2}{*}{CKA$_{Ref}$} \\
    & \proposal & & No Pruned Model &  \\
    \midrule
    PAINTING - & \multirow{2}{*}{0.002} & \multirow{2}{*}{0.992} & \multirow{2}{*}{0.471} & \multirow{2}{*}{0.259} \\
    Sculpture  &  &  &  &  \\
    \midrule
    PAINTING -  & \multirow{2}{*}{0.003} & \multirow{2}{*}{0.986} & \multirow{2}{*}{0.502} & \multirow{2}{*}{0.202} \\
    Painting    &  &  &  &  \\
    \midrule
    PAINTING -   & \multirow{2}{*}{0.002} & \multirow{2}{*}{0.993} & \multirow{2}{*}{0.433} & \multirow{2}{*}{0.236} \\
    Iconography  &  &  &  &  \\
    \midrule
    PAINTING - &  \multirow{2}{*}{0.006} & \multirow{2}{*}{0.984} & \multirow{2}{*}{0.441} & \multirow{2}{*}{0.291} \\
    Engraving    &  &  &  &  \\
    \midrule
    PAINTING - &  \multirow{2}{*}{0.021} & \multirow{2}{*}{0.925} & \multirow{2}{*}{0.432} & \multirow{2}{*}{0.236} \\
    Drawings    &  &  &  &  \\
    \midrule    
\end{tabular}
%}}
\end{table}

Table \ref{tab:painting-cka-both} shows the results for pruning consecutive layers in the dataset PAINTING. These results, again, prove that the closest and best elements of \proposal achieve a great value of CKA given a low value of Hamming distance, which is the best output that we can have. Moreover, the CKA values for the reference models are high, but this is due to the fact that the difference in accuracy between the models is lower. However, this also proves that \proposal also learns the best neurons for this problem.

\begin{table}[H]
\centering
\caption{\begin{idea}Comparison of the CKA measure for pruning consecutive layers in PAINTING-Class.\end{idea}}
\label{tab:painting-cka-both}
\resizebox{\columnwidth}{!}{%
{
\begin{tabular}{*{1}{c}*{5}{c}}
  \toprule
  % Cambiar orden del nombre
      &  & \multicolumn{3}{c}{\textbf{Pruning consecutive layers}}\\
    \cmidrule(lr){2-6}
    \multirow{2}{*}{Dataset} & \multirow{2}{*}{\# Layer} & \multirow{1}{*}{Hamming Distance} & \multirow{2}{*}{CKA$_{Closest}$}  & \multirow{1}{*}{Hamming Distance} & \multirow{2}{*}{CKA$_{Ref}$} \\
    & & \proposal & & No Pruned Model &  \\
    \midrule
    PAINTING -  & Layer 1 & \multirow{2}{*}{0.001} & 0.998 & \multirow{2}{*}{0.359} & 0.832 \\
    Sculpture   & Layer 2 &  & 0.999 &  &  0.879 \\
    \midrule
    PAINTING -  & Layer 1 & \multirow{2}{*}{0.004} & 0.998 & \multirow{2}{*}{0.578} & 0.701 \\
    Painting    & Layer 2 &  & 0.998 &  & 0.777 \\
    \midrule
    PAINTING -  & Layer 1 & \multirow{2}{*}{0.001} & 0.998 & \multirow{2}{*}{0.701} & 0,593  \\
    Iconography & Layer 2 &  & 0.998 &  & 0.672 \\
    \midrule
    PAINTING -  & Layer 1 & \multirow{2}{*}{0.001} & 0.999 & \multirow{2}{*}{0.255} & 0.895 \\
    Engraving   & Layer 2 &  & 0.999 & & 0.931 \\
    \midrule
    PAINTING -  & Layer 1 & \multirow{2}{*}{0.003} & 0.998 & \multirow{2}{*}{0.427} & 0.797 \\
    Drawings    & Layer 2 &  & 0.998 & & 0.874 \\
    \midrule    
\end{tabular}
}}
\end{table}

In this section, we have shown that \proposal is able to capture the relevance of the diverse classes and datasets that they are shown. Thanks to that adaption, \proposal has shown its robustness and its ability to search for the best neurons to tackle the problem at hand.

\subsubsection{Effects of a gradual aggregation of a class in the problem at hand}

This section is devised to analyze the impact of a class when it appears as a new class in a dataset, and it increases its number of examples over time. We have selected the class Iconography of PAINTING for these experiments. The reason lies in the fact that the class with a low percentage of examples is a minority class of this dataset, but it becomes the majority class of PAINTING when all the examples are used. Adding more examples of this class lets us check how \proposal is able to adapt to different scenarios for the same dataset when a class is gradually growing on its importance in the dataset.

This section has a similar structure to the previous one. First, we show the results of the experiments for pruning consecutive layers and feature selection models of \proposal. Then, the CKA values is also presented to perform the same double comparison as it has been done in the last section. For this section, the notation for the dataset is PAINTING-Iconography we talk about the dataset resulting from adding more examples, and in the tables this is shown as PAINTING-$Pct\%$, where $Pct = 20, 40, 60 \text{ and } 80$.

The first experiments we show are in Table \ref{tab:painting-icono}. These experiments have been carried out with the dataset PAINTING with the different percentage, as we have previously explained. The results show that, as the percentage of data increases, the models tend to become better. If we compare these results with those obtained with the full data set, we see that the model with 80\% of the data (and with 60\% of the data also for feature selection) is the closest to the full model (see Table \ref{tab:comparison-reference-consecutive} and Table \ref{tab:fs-ga}).

Reviewing the results tables with the full dataset and comparing them with these results, we observe that for the consecutive layer pruning model, the number of active neurons is lower in these experiments. The same phenomenon occurs in most of the feature selection cases, except when 40\% of the data is used, where this number increases as the model improves the accuracy for that dataset at the cost of increasing the percentage of active neurons.

% añadir las tablas de PAINTING con la clase iconography para el 20%, 40%, 60% y 80% de dicha clase
\begin{table}[H]
\centering
\caption{\begin{idea}Average results for PAINTING-Iconography with pruning consecutive layers and feature selection.\end{idea}}
\label{tab:painting-icono}
%\resizebox{\columnwidth}{!}{%
%{
\begin{tabular}{*{1}{c}*{5}{c}}
  \toprule
  % Cambiar orden del nombre
\multirow{2}{*}{Dataset} & \multirow{2}{*}{Measure} & \multirow{1}{*}{\proposal} & No Pruning  & \multirow{1}{*}{\proposal} & No Pruning \\
    & & \multirow{1}{*}{Both} & Both & \multirow{1}{*}{Feature Selection} & Feature Selection \\
    \midrule
    PAINTING - & Accuracy & 0.944 & 0.931 & 0.945 & 0.826 \\ 
    20\%  & \% Active neur. & 47 & 100 & 56 & 100 \\ 
    \midrule
    PAINTING - & Accuracy & 0.945 & 0.929 & 0.950 & 0.826  \\ 
    40\%   & \% Active neur.& 39 & 100 & 65 & 100 \\ 
    \midrule
    PAINTING -  & Accuracy & 0.946 & 0.931 & 0.954 & 0.839 \\ 
    60\% & \% Active neur. & 53 & 100 & 51 & 100 \\ 
    \midrule
    PAINTING - & Accuracy & 0.947 & 0.927 & 0.953 & 0.839 \\ 
    80\%  & \% Active neur. & 41 & 100 & 47 & 100 \\ 
    \midrule    
    PAINTING & Accuracy & 0.953 & 0.939  & 0.958 & 0.939 \\ 
    Full     & \% Active neur. & 51 & 100 & 55 & 100 \\  
    \midrule
\end{tabular}
%}}
\end{table}
    
Next, we follow the same process as for the previous section, in which the different classes of various data sets were analyzed. We show the value of CKA for the feature selection model and for the pruning consecutive layers model. Table \ref{tab:painting-icono-cka} shows the results for the different feature selection models applied to the various data percentage options of the Iconography class. The CKA value, which compares the best with its closest element in the history of the execution, is very high. This implies that \proposal is a robust model, since the phenotype obtained from the genotype is very similar. In addition, the other CKA value reported by the reference model indicates that \proposal is capable of the neurons important for the model, thus explaining the difference in accuracy between the two approaches.

% añadir las tablas de CKA para cada uno de los experimentos
\begin{table}[H]
\centering
\caption{\begin{idea}Comparison of the CKA measure for feature selection in PAINTING-Iconography.\end{idea}}
\label{tab:painting-icono-cka}
%\resizebox{\columnwidth}{!}{%
%{
\begin{tabular}{*{1}{c}*{4}{c}}
  \toprule
  % Cambiar orden del nombre
    &  \multicolumn{4}{c}{\textbf{Feature Selection}}\\
    \cmidrule(lr){2-5}
    \multirow{2}{*}{Dataset} & \multirow{1}{*}{Hamming Distance} & \multirow{2}{*}{CKA$_{Closest}$}  & \multirow{1}{*}{Hamming Distance} & \multirow{2}{*}{CKA$_{Ref}$} \\
    & \proposal & & No Pruned Model &  \\
    \midrule
    PAINTING - & \multirow{2}{*}{0.001} & \multirow{2}{*}{0.995} & \multirow{2}{*}{0.442} & \multirow{2}{*}{0.277} \\
    20\%  &  &  &  &  \\
    \midrule
    PAINTING -  & \multirow{2}{*}{0.007} & \multirow{2}{*}{0.976} & \multirow{2}{*}{0.352} & \multirow{2}{*}{0.461} \\
    40\%    &  &  &  &  \\
    \midrule
    PAINTING -   & \multirow{2}{*}{0.019} & \multirow{2}{*}{0.934} & \multirow{2}{*}{0.484} & \multirow{2}{*}{0.205} \\
    60\%  &  &  &  &  \\
    \midrule
    PAINTING - &  \multirow{2}{*}{0.001} & \multirow{2}{*}{0.998} & \multirow{2}{*}{0.530} & \multirow{2}{*}{0.177} \\
    80\%    &  &  &  &  \\
    \midrule
\end{tabular}
%}}
\end{table}

The following table, Table \ref{tab:painting-icono-cka-both}. The robustness of the proposal becomes evident when comparing the best element with its closest element in each of the runs, which has also occurred in the previous case. CKA values are extremely high when these elements are compared. 

The comparison with respect to the reference models shows results similar to those of other cases of consecutive layer pruning. The difference in accuracy is reflected in the CKA, which is higher than in the feature selection cases, because this difference is larger when it comes to the pruning of the extracted features of the network. 

\begin{table}[H]
\centering
\caption{\begin{idea}Comparison of the CKA measure for pruning consecutive layers in PAINTING-Iconography.\end{idea}}
\label{tab:painting-icono-cka-both}
\resizebox{\columnwidth}{!}{%
{
\begin{tabular}{*{1}{c}*{5}{c}}
  \toprule
  % Cambiar orden del nombre
    &  & \multicolumn{3}{c}{\textbf{Pruning consecutive layers}}\\
    \cmidrule(lr){2-6}
    \multirow{2}{*}{Dataset} & \multirow{2}{*}{\# Layer} & \multirow{1}{*}{Hamming Distance} & \multirow{2}{*}{CKA$_{Closest}$}  & \multirow{1}{*}{Hamming Distance} & \multirow{2}{*}{CKA$_{Ref}$} \\
    & & \proposal & & No Pruned Model &  \\
    \midrule
    PAINTING -  & Layer 1 & \multirow{2}{*}{0.001} & 0.999 & \multirow{2}{*}{0.529} & 0.732 \\
    20\%      & Layer 2 &  & 0.998 &  &  0.794 \\
    \midrule
    PAINTING -  & Layer 1 & \multirow{2}{*}{0.001} & 0.998 & \multirow{2}{*}{0.607} & 0.677 \\
    40\%    & Layer 2 &  & 0.999 & & 0.755 \\
    \midrule
    PAINTING -  & Layer 1 & \multirow{2}{*}{0.001} & 0.999 & \multirow{2}{*}{0.466} & 0.758 \\
    60\%    & Layer 2 &  & 0.999 &  & 0.828 \\
    \midrule
    PAINTING -  & Layer 1 & \multirow{2}{*}{0.001} & 0.998 & \multirow{2}{*}{0.594} & 0.694 \\
    80\%      & Layer 2 &  & 0.998 & & 0.762 \\
    \midrule
\end{tabular}
}}
\end{table}

This section has allowed us to see \proposal in different situations it has had to face. From data sets with fewer classes so that our proposal is able to adapt to all the subclasses that compose it (first subsection), to the gradual increase of a class from being the minority to the majority (second subsection). The study which has been performed in this section relies on a measure which allows us to study the robustness of \proposal and, in addition, allows us to see the differences in accuracy of the models that they have been developed.

The results in both sets of experiments show that the stochasticity that might be present in the proposal is not influential. The results of the CKA measure when comparing the best trained network found by \proposal and its closest trained network for each of the runs show the high degree of robustness of \proposal.

The comparison of \proposal with networks with all neurons active (CKA$_{Ref}$) shows us a twofold conclusion. When we are dealing with the feature selection models, those that performed the pruning of the extracted features of the network, the difference in accuracy is reflected in the value of CKA, which is very low and that means that the models are very different On the other hand, for the case of pruning consecutive layers, the CKA$_{Ref}$ value reflects models with less difference in comparison to the previous case. However, both in feature selection and pruning consecutive layers it is observed that \proposal is able to search for the neurons that best approximate the problem to be solved.
\end{idea}

\begin{idea}
\section{Advantages and disadvantages of \proposal}\label{sec:lessons}

This section is devised to discuss the advantages and disadvantages of \proposal, considering the diverse and large experimentation which has been done in the manuscript.  The advantages of \proposal can be summarized as follows:

\begin{itemize}
    \item \textbf{Specialization of the last layers of networks.}
    
    An important element of \proposal is the transfer learning. This is one of the most commonly used techniques. We have refined its process, which is the extraction of pre-trained features and then, the specialization of the fully-connected layers. In this context, \proposal, and specifically the GA which is composed of, when applied to these layers does not  limit the network learning compared to other evolutionary models in the literature that require high computational time to evaluate the datasets, as they train the whole network. For that reason, \proposal can be applied to more complex datasets.
    
    \item \textbf{Performance over reference models and efficient pruning methods from the literature.}
    
    The usage of an evolutionary model that focuses on pruning neurons of the fully-connected layers achieves a better performance than other pruning methods when applied under the same conditions of \proposal. The positive effect of the genetic algorithm is the selection of the best neurons of these layers, so that the evolution towards the best configuration for the networks is obtained thanks to \proposal.
    
    \item \textbf{Constructive modeling over the last layers of the networks.}
    
    In the different experiments that have been carried out in the sections, we have observed that performing the pruning constructively based on the number of layers achieves good results. Pruning one-layer networks achieves good results, but when the number of layer increases, it is shown that performing the pruning over a single layer of two-layer networks improves the one-layer networks. Nonetheless, the simultaneous pruning of the both layers achieves a better modeling of the datasets than all the previous pruning models.
    
    \item \textbf{Pruning the extracted features of the network against pruning fully-connected layers.}
    
    \proposal obtains better results by pruning the self-generated features resulting from transfer learning versus pruning the fully-connected layers. This is an intuitive idea because the learned patterns or features are different for each problem, and the learned features for the original problem may not be useful for the target problem. Knowing which characteristics matter is crucial to the problem at hand. The evolutionary process allows pruning these features and selecting those that best solve the modeling problem under consideration.
    
    \item \textbf{Generalization of \proposal to other feature extractors.}
    
    One of the advantages of \proposal is that the model is generalizable to diverse feature extractors. This is an important advantage because \proposal is able to achieve a great performance over different datasets and with diverse networks. These are used to extract the features of the dataset, thanks to the transfer learning technique. \proposal has improved the reference models and pruning models from the literature in both the pruning of two consecutive layers and the pruning of the features extracted by the networks.

    \item \textbf{Adaptation to relevant classes and gradual aggregation of data for the problem at hand.}
    
    \proposal has achieved a great performance in different situations. However, it is also important to see how it adapts to different situations within the datasets themselves. \proposal has improved the performance of the reference models, but a measure is needed to support the quality of these models. However, this increase in performance has been proven by CKA not to be the result of chance, so \proposal models are able to search for the best neurons to maximize the accuracy of the problem.
    
    \item \textbf{The good results of \proposal do not depend on randomness.}
    
    The evolutionary algorithm in which \proposal lies in is a stochastic algorithm and the results may be affected by randomness. This is a risk about the model, so it is required to check if the good results are biased by the randomness. For that reason, we have introduced a new measure, CKA. This measure compares the differences between the pruned networks, and it is a way to measure the robustness of \proposal. The values of the CKA for the various experiments of the previous section shed light on the fact that \proposal results do not depend on the randomness. 

\end{itemize}

The main disadvantage of \proposal is:

\begin{itemize}
    \item \textbf{Execution time of \proposal.} 
    
    The main drawback of \proposal it is the time that is required to execute the model. In comparison with the pruning methods from the literature and the reference models, the table of execution times (see Table \ref{tab:times-evodeeptlpruning-literature}) shows the speed of the other models, but \proposal is slower than these models. The time difference is made up by improved network performance, thanks to the usage of \proposal. Nonetheless, there are some practical cases in which the training time is not a problem, like in medical diagnosis, because the main objective is obtaining a better percentage of the models in terms of accuracy.
\end{itemize}

\end{idea}

\section{Conclusions} \label{sec:conclusions}

This paper has introduced \proposal, a novel model that sparsifies the architecture of the last layers of a DL model initialized using TL. \proposal is a combination of sparse layers and EA, so that the neurons of these layers are pruned using the EA, in order to adapt them to the problem to tackle and deciding which neurons/connections to leave active or inactive. 

\proposal is a flexible model that \begin{idea2}evolves\end{idea2} models with one and two layers and even two layers at the same time. Our results show that the pruning over complete neurons is better than pruning connections individually, establishing the last one as the best encoding strategy. The evolution of the sparse layer improves these models in terms of accuracy and also in terms of complexity of the network. In comparison with compared reference models and pruning methods from the literature, \proposal achieves a better performance than all of them. The choice of one among the pruning models or feature selection has been answered and informed with experimental evidence: the FS scheme derived from \proposal has shown a better performance, in most cases, than the pruning methods\begin{idea}. The ability of adaptation of \proposal to other feature extractors has been tested. Lastly, \proposal has also shown its capability to adapt to the relevance of diverse problems and it has also achieved an outstanding level of robustness, which implies that the results do not depend on random nature of the search operators used by the GA that lies at the core of the proposed evolutionary pruning method. \end{idea}

From an overarching perspective, this work aligns with a growing strand of contributions where evolutionary computation and DL have synergized together to yield \begin{idea2}evolved\end{idea2} models that attain better levels of performance and/or an increased computational efficiency. Indeed, this fusion of concepts (forged as Evolutionary Deep Learning) has been used for other \begin{idea2}evolution\end{idea2} processes, including hyperparameter or structural tuning. Another recent case of the symbiosis of EA's and DL are represented in AutoML-Zero that use an evolutionary search to automatically search the best DL structure. AutoML-Zero and \proposal are two great examples of the benefits of combining EA's and DL that outline the potential and promising path of successes envisioned for this research area.

Future research work stemming from the results reported in this study is planned from a two-fold perspective. To begin with, we plan to achieve larger gains from the combination of DL and EA by extending the evolutionary search over higher layers of the neural hierarchy, increasing the number of \begin{idea2}evolved\end{idea2} layers and neurons per layer. To this end, we envision that exploiting the layered arrangement in which neurons are deployed along the neural architecture will be essential to ensure an efficient search. The second research line relates to this last thought, aiming to improve the search algorithm itself by resorting \begin{idea}to \end{idea}advanced concepts in evolutionary computation (e.g. niching methods or co-evolutionary algorithms).

\begin{idea2}
Finally, a third research path arising from this work is the reformulation of the pruning problem and the adaptation of EvoPruneDeepTL to select which activation functions to be used in the fully connected part of the neural network. Our hypothesis is that the pruning problem can be reformulated so that every decision variable represents which activation function to utilize in every neuron. Efforts will be invested in this direction, adapting operators within EvoPruneDeepTL to efficiently explore the combinatorial search space of the reformulated optimization problem.    
\end{idea2}

%A double perspective comes in the future works to take advantage of DL and EA. The first perspective refers to DL. The optimization of the network in terms of number of layers and neurons per layer offers a more explainable network with an increase of its performance. The second perspective is related to the evolutionary scheme of our model. The usage of more advanced techniques like collaborative populations or even niching methods will be useful to develop models with even better performance.

%In future works, we will study the usage of more efficient architectures with the intention to achieve better results in less computational time. Apart from that, the combination of EA and DL has proven the benefits of their synergy. 
%We will consider the optimization of the number of layers and the number of neurons per layer to achieve a more detailed optimization of the network. Moreover, we will also study the usage of more complex models such as niching methods for EA to develop models based on the interaction of the best configurations to get models with better performance.

\section*{Acknowledgments}

F. Herrera, D. Molina and J. Poyatos are supported by the Andalusian Excellence project P18-FR-4961, the infrastructure project with reference EQC2018-005084-P and the R\&D and Innovation project with reference PID2020-119478GB-I00 granted by the Spain's Ministry of Science and Innovation and European Regional Development Fund (ERDF). Aritz D. Martinez and Javier Del Ser would like to thank the Basque Government for the funding support received through the EMAITEK and ELKARTEK programs, as well as the Consolidated Research Group MATHMODE (IT1456-22) granted by the Department of Education of this institution.

\bibliographystyle{model5-names}

%\bibliographystyle{cas-model2-names}

\bibliography{bibliography}

\end{document}